\pdfoutput=1

\documentclass[11pt]{article}
\usepackage[table,xcdraw]{xcolor}
\usepackage[]{acl}

\usepackage{times}
\usepackage{xcolor}
\usepackage{latexsym}
\usepackage{soul}

\usepackage{enumitem}
\usepackage{wrapfig}
\usepackage{bbm}
\usepackage{array}
\usepackage{subcaption}
\usepackage{graphicx}

\def \cY {\cal Y}
\def \cX {\cal X}

\def \x {\mathbf{x}}
\def \y {\mathbf{y}}
\def \z {\mathbf{z}}

\newcommand{\E}{\mathbb{E}}
\newcommand{\D}{\mathcal{D}}

\usepackage[normalem]{ulem}
\useunder{\uline}{\ul}{}
\setlist{nosep}
\usepackage[T1]{fontenc}

\usepackage[utf8]{inputenc}

\usepackage{microtype}

\usepackage{enumitem}
\setlist{nosep}
\usepackage{hyperref}
\usepackage{url}
\usepackage{caption}
\usepackage{subcaption}
\usepackage{graphicx}
\usepackage{graphics}
\usepackage{amssymb}
\usepackage{amsmath}
\usepackage[utf8]{inputenc} 
\usepackage[T1]{fontenc}    
\usepackage{hyperref}       
\usepackage{url}            
\usepackage{booktabs}       
\usepackage{amsfonts}       
\usepackage{nicefrac}       
\usepackage{microtype}      
\usepackage{xcolor}         
\usepackage{array}
\usepackage{booktabs,caption}
\usepackage[flushleft]{threeparttable}
\usepackage{authblk}
\newcommand{\PreserveBackslash}[1]{\let\temp=\\#1\let\\=\temp}
\newcolumntype{C}[1]{>{\PreserveBackslash\centering}p{#1}}
\newcolumntype{R}[1]{>{\PreserveBackslash\raggedleft}p{#1}}
\newcolumntype{L}[1]{>{\PreserveBackslash\raggedright}p{#1}}

\title{Filtering Context Mitigates Scarcity and Selection Bias in Political Ideology Prediction}

\author[1]{Chen Chen\thanks{chenchen2020@cuhk.edu.cn}}
\author[2]{Dylan Walker\thanks{Dylan@chapman.edu}}
\author[3]{Venkatesh Saligrama\thanks{srv@bu.edu}}

\affil[1]{School of Management and Economics, Chinese University of Hongkong (Shenzhen)}
\affil[2]{Argyros School of Business and Economics, Chapman University}
\affil[3]{College of Engineering, Boston University}

\begin{document}
\maketitle
\setlength{\abovedisplayskip}{3pt}
\setlength{\belowdisplayskip}{3pt}

\begin{abstract}

We propose a novel supervised learning approach for political ideology prediction (PIP) that is capable of predicting out-of-distribution inputs. This problem is motivated by the fact that manual data-labeling is expensive, while self-reported labels are often scarce and exhibit significant selection bias. We propose a novel statistical model that decomposes the document embeddings into a linear superposition of two vectors; a latent neutral \emph{context} vector independent of ideology, and a latent \emph{position} vector aligned with ideology. We train an end-to-end model that has intermediate contextual and positional vectors as outputs. At deployment time, our model predicts labels for input documents by exclusively leveraging the predicted positional vectors. On two benchmark datasets we show that our model is capable of outputting predictions even when trained with as little as 5\% biased data, and is significantly more accurate than the state-of-the-art. Through crowd-sourcing we validate the neutrality of contextual vectors, and show that context filtering results in ideological concentration, allowing for prediction on out-of-distribution examples. 

\end{abstract}

\section{Introduction} \label{sec:intro}
Political Ideology Prediction (PIP) is motivated by a number of applications such as the need to understand government's policy making \citep{potrafke2018government, hunt2009ideology}, a legislator's partisan/non-partisan actions \citep{zingher2014ideological, nice1985state, berry2007measurement}, the general public's sentiment support for legislation \citep{berry2007measurement,budge1987ideology,rudolph2009political} etc.

Our overall goal is to infer ideology of legislators or the general public 
from written texts, media posts, and speeches. We can train a PIP model with ground truth labels (e.g., liberal vs conservative) with a standard supervised learning approach. In some cases, such as for legislators, ground-truth labels can be inferred through party affiliation information, and these have served as good proxies for ideology \cite{poole2001d}.

\noindent {\bf Selection Bias.} In other situations, such as, for the domain of social media posts, ground truth is difficult to obtain, and is based on self-reported affiliation of the users \citep{bakshy2015exposure}. Such self-reported affiliations are generally sparse, and when reported tend to be of extreme polarity \citep{conover2011predicting}. While manual labeling by annotators (e.g., AMTs) \citep{preoctiuc2017beyond} can be leveraged, selection bias is still an issue due to oversampling of posts from extreme vocal posters \citep{cohen2013classifying, mustafaraj2011vocal,kohut2018vocal, gao2015vocal}.
%

\noindent {\bf Justification.} A black-box model trained exclusively on scarce and polarized group is likely to perform poorly on the under-observed moderates who are the majority and are reticent to disclose or discuss their politics online \cite{pewstudy,kohut2018vocal,gao2015vocal}. 
Inferring the majority's views is important to help us understand support for legislative/executive actions \citep{craig2014precipice}, or cultural group's real influence \citep{bisin2000model} as in the recent Kansas vote where policymakers over-estimated the public support for the abortion ban by overlooking the voice of silent/less vocal majority\citep{nprnews}. 



\noindent {\bf Proposed Method.} To account for scarcity and selection bias, we propose a novel method that, during training, enforces decomposition of text embedding into context and position, and trains a deep neural network model in a modified variational auto-encoder framework with bi-modal priors. 


\noindent {\bf Contextual Filtering.}
We propose to decompose document embeddings into a linear (orthogonal) superposition of two {\it latent} vectors consisting of a context vector and a position vector. We train a DNN model embedded with this decomposition to enforce the fact that context vectors be devoid of information that contains ideology, and the residual position vectors, obtained by filtering out the context, exclusively bears all ideological information. To ensure that trivial solutions are not found, we require that context vectors across all documents belong to a low-dimensional manifold. Our perspective is that there are a sparse collection of common words (eg. guns, immigration, taxes, etc) whose neutral components serve to contextualize the document, and their corresponding embeddings constitute a low-dimensional space. 

\noindent {\bf De-Noising.}
Documents exhibit significant variance in scarce regimes, making it difficult to discern ideology. This is due to scarce document collections spread across diverse themes. Position vectors, devoid of context, suppresses noise that accounts for large differences between document embeddings, but carries little relevance for PIP, making learning representation of ideology easier.



\noindent {\bf Key Experimental Findings}\\
\underline{SOTA Prediction.} Ideology Prediction by our  model BBBG across scarce and biased labels regimes was universally dominant with respect to prior works. 
Furthermore, compared to other embeddings such as BERT/RoBERTa, GLoVe predictions were more accurate particularly in scarce/biased regimes.\\
\underline{Neutrality of Latent Contexts.} Crowd-sourced experiments showed context vectors are associated with neutral words/phrases across themes.\\
\underline{Ideological Purity.} Context Filtering leads to ideological purification and improves prediction.\\
\underline{Knowledge Transfer.} BBBG was able to effectively transfer knowledge to out-of-domain scenarios such as (a) generalizing ideology prediction from extremely polarized authors to near-moderate authors; (b) generalizing documents from ``seen themes'' to documents from ``unseen themes.''

\section{Related Work} \label{sec:related}
We describe prior works that focus on machine learning for predicting ideology from texts.

\noindent {\it Supervised Learning.} \cite{gentzkow2010drives} propose a learning-based approach for ideology prediction on congressional report corpora. Other studies on the same dataset apply modern learning methods under the same setting \citep{pla2014political,blei,iyyer2014political}. Research using social media data (tweets, forum posts) aims to map public users onto the liberal-conservative spectrum or simply predict their party labels \citep{levendusky2010measuring,baldassarri2008partisans}. \\ 
\noindent {\it Textual features.}
The majority of the aforementioned studies utilize off-the-shelf pretrained text representations such as BOW, TF-IDF, LIWC, or Word Embeddings--GloVE, BERT or RoBERTa \citep{preoctiuc2017beyond, conover2011predicting, mokhberian2020moral, liu2019roberta}. 

\noindent{\it Scarce and Biased Data.}
The labels used as supervision are obtained from self-reported party affiliations (e.g., Democrats or Republicans) 
or manual labeling by annotators \citep{conover2011predicting,preoctiuc2017beyond,cohen2013classifying}, and inherently suffer from label scarcity and selection bias. For example, the majority of the public are reticent to disclose their party affiliation or engage in political discourse online \citep{bakshy2015exposure,pewstudy}. Furthermore, methods proposing to collect labeled texts to study opinions or ideology are prone to over-sampling the "vocal minority" while ignoring or down-sampling the "less vocal majority" \citep{moe2011influences,kohut2018vocal,mustafaraj2011vocal,gao2015vocal}. 
Manual labeling is severely constrained by staffing and cognitive biases of the annotators themselves \citep{yu2008classifying, xiao2020timme}. Finally, there are large domain differences between content generated by visibly opinionated and  moderate users \citep{cohen2013classifying}. For various reasons, models for learning ideology typically suffer from poor generalizablity \citep{cohen2013classifying, yan2017perils,yan2019congressional}. 

\noindent {\it General Methods.} So far there is little research that jointly accounts for both label scarcity and selection bias in training ideology prediction models based on textual data. For Twitter data, social interaction of users (eg. mentions/follows) are leveraged in a graph learning framework to combat the label scarcity issue. However, these methods may not apply to settings where social interaction and connection are absent or unobserved \citep{xiao2020timme}. Traditionally, semi-supervised learning (SSL) can deal with insufficient labels (see \citep{ortigosa2012sslsentiment,oliver2018realistic,kingma2014semi}), whereas Domain Adaptation (DA) \citep{daume2009frustratingly,glorot2011domain} can deal with the distribution discrepancy between training and test samples, potentially providing a solution to selection bias. We compare these alternative approaches and word-embeddings, and other prior works in our experiments. 

\section{Statistical Model of Text Generation} \label{sec:decomp}


\noindent \textbf{Document Decomposition.} 
We first encode documents in a suitable embedding space (GloVE, RoBERTa etc.), and let $\x \in \cX \subset \mathbb{R}^D$ represent this embedding, which serves as inputs to the learner. 
We decompose, $\x$ into two major components: 
\begin{equation}
 \mathbf{x}= \mathbf{c}(\theta) + \mathbf{f}(\theta, z) + \mathbf{\epsilon}
 \label{eq:decomp}
\end{equation}
where 
$~\mathbf{z}\in \mathbb{R}^M$ is the latent ideology vector for $~\mathbf{x}$, and vectors $~\mathbf{c}$ and $~\mathbf{f}$ denote the neutral context and filtered position vector components, respectively.
$\mathbf{\epsilon}$ is a random vector that captures idiosyncratic variations among documents. The parameter $~\mathbf{\theta}$ represents the author's choice of themes such as "guns" or "abortion". However, such decomposition is unidentifiable without further constraints. We impose a low-dimensional structure on context vectors, and during training disambiguate context vectors from ideological content. 

\noindent\textbf{Low-Dimensional and Neutral Encoding of Context Vectors.} 
We proceed with the following intuition. For a certain choice of partisan theme $\theta_i$, say that there is a major \textit{p}olarization \textit{a}xis $pa_{\theta_i}$ depending on the theme, where people of different ideological groups (such as liberal or conservative, left or right) differentiate when giving speeches\footnote{Mathematically, this can be approximated either via the first principal axis or, in a simpler way, differentiating all average embeddings from the liberal group over the conservative group.}. In principle we propose to seek orthogonality of context to the polarization axis $pa$. In practice, since $pa$ is unobserved \textit{ex-ante}, we enforce this constraint by careful initialization and empirically aligning position $\mathbf{f}$ with polarity during training, and we verify the orthogonality \textit{ex-post}. 

\noindent \textit{Initialization.} To initialize context vectors, we adopt the following procedure. We first determine themes by generating a set of "neutral" seed words and phrases (hereafter, seeds), either through manual crafting or selecting frequent words with low $\chi^2$ values (see Appendix Sec.~\ref{sec:initialization}), and complement this set by expanding into the neighborhood of seeds, yielding a set of seeds for each theme $i$, $a_{ij}$. Themes are initialized as the pooled embedding of seed words $T_{0i}=\sum_j a_{ij}$. Context vectors are initialized as a mixture of themes, $c_0(\theta_0)=T_0 \theta_0$. Ultimately, both the theme choice, $\theta$ and the themes $T$ are learned through training, starting with the initialization $T=T_0$ (more details and alternate initializations are in the Appendix A.4 and A.5) and low-dimensionality results from the fact that the theme matrix is low-rank. This approach effectively removes variance that accounts for large differences between document embeddings, but carries little relevance for ideology learning and prediction, hence making learning representation of ideology easier. 

\noindent \textbf{Multimodal Prior.} Because it is commonplace for individuals to explicitly identify (partisanship) with groups of common ideological position, we presume that representation of ideology, $\mathbf{z}$ is drawn from a multi-modal prior. A reasonable choice of modality in the domain of the US politics is two, indicating bipartisanship/polarization. In the context of the congress and political debate forums, this assumption is supported by previous studies \citep{poole2001d,bonica2014mapping} and reflects a general trend in the US society \footnote{https://www.pewresearch.org/politics/2014/06/12/\newline political-polarization-in-the-american-public/}. However, our framework can generalize to multiple modes.

\section{Method} \label{sec:method}
\noindent \textbf{Mathematical Background} 
In addition to notation in Sec.~\ref{sec:decomp}, we let $\y \in \cY$ denote the output ideological labels taking values in a finite set. 
Let $p_s(\x,\y), p_t(\x,\y)$, denote source and target joint distributions respectively (we drop these subscripts when clear from context). The learner gets access to a labeled set, $\D_s$ of $N_s$ \textit{i.i.d} data points $(\x_i,\y_i)\sim p_s$ and unlabeled set, $\D_t$, of $N_t$ \textit{i.i.d} target input instances $\x_i \sim p_t$. 

\noindent \textbf{Inputs and Outputs of Model.} The goal of the learner is to predict labels $\mathbf{\hat y}$ for the target given target inputs. Additionally, our trained model also outputs for each input document, the context vector, $\mathbf{c}$, the theme choices, $\theta$, the position vector, $\mathbf{f}$, the ideology vector, $\mathbf{z}$. 

\noindent \textbf{Training Loss.} Our
loss function is a sum of reconstruction loss for inputs $\x_i \in \D_{s}\cup \D_{t}$, and cross-entropy loss, $CE(\y_i,\mathbf{\hat y}(\z_i;\gamma))$ on $(\x_i,\y_i) \in \D_{s}$; $\mathbf{\hat y}$ is the softmax output of a network governed by parameters $\gamma$, and taking the encoded latent representation $\mathbf{z}_i$ (see below) as its input. 

\noindent \textit{Reconstruction Loss.} 
Our starting point is to optimize the marginal distribution, namely, $\E_{\x \sim q(\x)}\log p(\x)$, but as this is intractable, we follow by relaxing it to the ELBO bound \citep{kingma2013auto}: 
\begin{equation}
    \label{VAE}
    \begin{split}
        L_R(\eta, \phi,\lambda) &\triangleq  \mathbb{E}_{\mathbf{x}\sim q(\mathbf{x})}[\mathbb{E}_{q_{\phi}(\mathbf{z}|\mathbf{x})}[\ln p_{\eta}(\mathbf{x}|\mathbf{z})\\
        &+ \ln p_{\lambda}(\mathbf{z}) - \ln q_{\phi}(\mathbf{z}|\mathbf{x})]] 
    \end{split}
\end{equation}
where, $q(\x)$ is the empirical distribution on both source and target inputs; $q_\phi(\z|\x)$ is the encoder, and $p_\eta(\x|\z)$ is the decoder; and $p_\lambda(\z)$ is the prior. $\phi, \eta, \lambda$ are their parameters respectively. We now invoke our statistical model Eq.~\ref{eq:decomp} to decompose $\mathbf{x}$ into $\mathbf{f}$ and  $\mathbf{c}$, and by neutrality of context vectors,  $p(\mathbf{c}| \mathbf{\theta, \mathbf{z}})= p(\mathbf{c}| \mathbf{\theta})$, and noting that conditioned on $\theta$, $\mathbf{c}$ and $\mathbf{f}$ are independent, we get,
\begin{align}
    \nonumber
        L_R(\eta, \phi,\lambda,\theta) &\triangleq  \mathbb{E}_{\mathbf{x}\sim q(\mathbf{x})}[\mathbb{E}_{q_{\phi}(\mathbf{z}|\mathbf{x})}[\ln p(\mathbf{f}|\mathbf{z}, \mathbf{\theta})]\\  &+\ln p(\mathbf{c}|\mathbf{\theta})+ \ln p_{\lambda}(\mathbf{z}) - \ln q_{\phi}(\mathbf{z}|\mathbf{x})\nonumber\\&+\ln p(\mathbf{\epsilon}|\mathbf{c} ,\mathbf{f})] \label{VAE_modified}
\end{align}

The first term in Eq. \ref{VAE_modified} corresponds to the decoder that generates $\mathbf{f}$; the second term describes the generation of $\mathbf{c}$ from $\mathbf{\theta}$. The third term $p_{\lambda}(\z)$ is the prior distribution of $\mathbf{z}$. The last term corresponds to minimizing reconstruction error($\mathbf{\epsilon} \triangleq \x - (\mathbf{f}+\mathbf{c})$) via mean square loss. Unlike traditional VAE, we enforce multi-modal prior and model it as a K-modal Gaussian following the approach in \citep{tomczak2018vae}. 
We approximate our prior by sampling and feeding $K$ pseudo-inputs into the same encoder, namely, 
$p_{\lambda}(\mathbf{z})=\frac{1}{K}\sum_{k=1}^{K}q_{\phi}(\mathbf{z}|\mathbf{\mu}_{n}).$
Here $K$ is a hyperparameter that specifies the modality of the prior. For the domain of the US politics we choose $K=2$, but our method can generalize to settings with different $K$. 

\noindent {\bf Deep Neural Network (DNN) Training.} 
We train a DNN by optimizing the total loss, $L=L_R(\eta,\phi,\lambda)+\mathbb{E}_{(\x,\y)\in {\cal D}_S} CE(\y,\mathbf{\hat y}(\z;\gamma))$ end-to-end by backpropagation over all the parameters $(\eta,\phi,\lambda,\gamma)$, while using a bi-modal prior on $\z$, and call the resulting predictor Bi-Branch Bi-Modal Gaussian Variational Auto-Encoder (BBBG). Our model differs from the traditional VAE in that it uses a multi-modal prior and reconstructs the input using two separately learned components in the generative parts of the model (see Appendix Fig~\ref{fig:dag},~\ref{fig:diagram}).  

\noindent \textit{The Single Branch Ablation.} As an ablation, and to understand the importance of Eq.~\ref{eq:decomp}, we relax the neutrality constraint by deleting the neutral context learning branch of the model and we call this algorithm SBBG (single-branch variant). In the simplified version, the contexts are no longer estimated and we are reduced to a standard VAE framework with a Gaussian mixture prior, but with the same $K$-modal VampPriors . Implementation and training details are described in Appendix B.2\\
%
%
\noindent \textit{Justification for Supervision.}
Traditional VAE models are unsupervised. But they tend to perform poorly due to over-regularization \citep{dilokthanakul2016gmvae}. In both of our experiments, the unsupervised VAE variant appears to collapse into a prior of uni-modal Gaussian. To solve this issue, we induced the model to converge to a bi-modal prior by providing supervision with a few labeled examples. This means that, during training, we allow \textit{some} ideology labels to be seen by the model, and the prediction loss of ideology are back-propagated to help tune the parameters in the encoders. The number of labels required to effectively train such a system turns out to be  5$\sim$8\% of the total samples. 
\vspace{-0.1in}

\section{Experiments} \label{sec:expt}
We experiment with two benchmark datasets, \textit{Congressional Speeches} and \textit{Gun Debate Forum} to baseline proposed BBBG, against well-known and prior ML methods. While we provide details of these datasets in the Sec.~\ref{sec:dataset_append}, we note that the reason for choosing these datasets is driven by the need to ascertain ground-truth extremity of texts and authors (ranging from extreme to neutral). Other datasets are binary, and as such this information is not provided. The datasets we chose allow for sub-sampling of sub-populations with varying levels of ideological extremity. This allows us to validate proposed method under selection bias. 

\noindent \textbf{Simulating Label Scarcity and Selection Bias.} 
We mask ground truth labels in a dataset to simulate label scarcity and selection bias. We define the level of supervision as the percentage of unmasked samples, $Sup$ which determines the extent of masking under either scarcity or scarcity and bias (toward extremity). To simulate scarcity, we masked $(1-Sup)\%$ of the data randomly. To simulate scarcity and selection bias, we masked $(1-Sup)\%$ of data from the least extreme authors. We refer to the former procedure as \textit{Unbiased Supervision} and the latter as \textit{Biased Supervision}. Under masking, the prediction loss from masked samples was not used in SGD to update weights of the network. We list details on the masking procedure, and tuning of hyperparameters in Sec. \ref{sec:methoddetails} ,~\ref{sec:addmask}.

\noindent {\bf Categorization By Themes.} 
We seek to expose the role of BBBG's other outputs, namely, the context vector, $\mathbf{c}$, the filtered position vector, $\mathbf{f}$, and the ideology vector, $\mathbf{z}$. To do so, for ablative purposes, we manually organize (note that BBBG training is agnostic to our categorization) Congressional dataset into 68 themes (see Sec.~\ref{sec:themes}) consisting of about 25 partisan and 43 non-partisan themes. We perform various experiments to study context neutrality, de-noising through context filtering, and knowledge transfer to unseen themes.

\noindent \textbf{Baseline Methods.} Prior methods leverage pre-trained text feature embeddings, and utilize supervised and unsupervised data in various ways. 
The GloVe embedding is the default embedding for both our main model (BBBG) and other baseline models. Models based on BERT or RoBERTa will be specified through naming. The complete implementation details and comparisons of all models appear in the Sec \ref{sec:methoddetails}.  

    \noindent {\it Methods trained only on labeled data.} These include GS \citep{gentzkow2010drives}, and standard ML methods such as SVMs, random forests (RFs), XGBoost, and 8-layer Deep Neural Networks (8l-DNN) with similar capacity as BBBG  and Gated Recurrent Units. These methods rely on pre-trained text features such as GloVe or RoBERTa \citep{dey2017gate, liu2019roberta}. \\
    \noindent {\it Methods leveraging both labeled and unlabeled target data.} These include semi-supervised learning (SSL) methods such as label-spreading with K-nearest neighbor (LS-KNN) \citep{ ortigosa2012approaching} as well as self-training (ST) methods combined with deep learning (ST-DNN) of similar capacity as BBBG and ensemble learning such as random forests (ST-RF)  \citep{yarowsky1995unsupervised,tanha2017semi,zhang2014semi}. ST is based on iterative pseudo-labeling \citep{zou2018unsupervised}; and finally Domain Adaptation (DA) methods that are built on RoBERTa embedding (RoBERTa-DA) are applied to handle domain shifts \citep{glorot2011domain}.  


\subsection{Prediction on Congressional Dataset}

\begin{table*}[t]
\caption{\footnotesize Accuracy of party prediction under unbiased/biased supervision for \textit{Congressional Speeches} data, showing competing results between the top three baselines, and the main model BBBG, and the single branch variant SBBG. The best results are in \textbf{bold} and the second best are \underline{underlined}. BBBG outperforms most other models substantially with scarce labels, marked in blue. The percentage shown were averaged over three independent trials.} \vspace{-0.1in}
\centering
\label{tab:congress}
\resizebox{16cm}{!}{%
\begin{tabular}{lccccccccccccccccc}
\hline
             & \multicolumn{8}{c}{\textbf{Unbiased Supervsion}}                                                                                                                                                                                                                                              &           & \multicolumn{8}{c}{\textbf{Biased Supervision}}                                                                                                                                                                                               \\ \cline{2-9} \cline{11-18} 
             & 80\%            & 60\%            & 40\%                                    & 20\%                                    & 8\%                                     & 5\%                                     & 3\%                                     & 1\%                                     &           & 80\%            & 60\%            & 40\%            & 20\%            & 8\%                                     & 5\%                                     & 3\%                                     & 1\%                                     \\
8l-DNN       & 93.2\%          & \textbf{94.4\%} & 92.6\%                                  & 88.1\%                                  & 84.9\%                                  & 55.0\%                                  & 53.6\%                                  & 50.4\%                                  &           & 84.5\%          & \textbf{84.8\%} & {\ul 86.9\%}    & \textbf{87.9\%} & 65.8\%                                  & 61.1\%                                  & 56.7\%                                  & 50.7\%                                  \\
GS           & 79.1\%          & 79.9\%          & 81.4\%                                  & 82.3\%                                  & 79.0\%                                  & 77.1\%                                  & 78.7\%                                  & 70.4\%                                  &           & 70.4\%          & 71.5\%          & 74.5\%          & 72.0\%          & 69.5\%                                  & 71.1\%                                  & 68.4\%                                  & 60.7\%                                  \\
LS-KNN       & 90.2\%          & 92.6\%          & 92.3\%                                  & 91.5\%                                  & {\ul 88.8\%}                            & 85.8\%                                  & 82.5\%                                  & {\ul 79.7\%}                            &           & {\ul 84.7\%}    & 83.5\%          & 83.0\%          & 81.4\%          & 78.2\%                                  & 75.7\%                                  & 71.3\%                                  & {\ul 67.8\%}                            \\
ST-DNN       & \textbf{93.5\%} & {\ul 94.3\%}    & 94.0\%                                  & {\ul 93.0\%}                            & 88.0\%                                  & 77.3\%                                  & 75.3\%                                  & 61.6\%                                  &           & 81.3\%          & 81.6\%          & 83.2\%          & 80.8\%          & 71.5\%                                  & 63.8\%                                  & 56.3\%                                  & 51.9\%                                  \\
RoBERTa-GRU  & 92.8\%          & 93.3\%          & 83.3\%                                  & 78.2\%                                  & 69.9\%                                  & 51.6\%                                  & 51.7\%                                  & 51.9\%                                  &           & 83.6\%          & 84.7\%          & 84.0\%          & 78.1\%          & 53.1\%                                  & 49.5\%                                  & 50.7\%                                  & 50.9\%                                  \\
RoBERTa-DA   & 90.6\%          & 89.1\%          & 89.4\%                                  & 79.3\%                                  & 75.1\%                                  & 62.9\%                                  & 61.8\%                                  & 51.4\%                                  &           & 84.0\%          & 81.0\%          & 81.9\%          & 68.5\%          & 69.3\%                                  & 54.8\%                                  & 51.4\%                                  & 49.8\%                                  \\
RoBERTa-SBBG & {\ul 93.3\%}    & 91.1\%          & {\ul 94.0\%}                            & 89.2\%                                  & 76.5\%                                  & 61.1\%                                  & 52.7\%                                  & 51.3\%                                  &           & \textbf{86.1\%} & {\ul 84.7\%}    & \textbf{87.5\%} & {\ul 85.9\%}    & 51.3\%                                  & 51.2\%                                  & 48.6\%                                  & 48.8\%                                  \\
SBBG         & 92.9\%          & 90.4\%          & 90.6\%                                  & 89.9\%                                  & 86.6\%                                  & {\ul 86.7\%}                            & {\ul 84.0\%}                            & 74.1\%                                  &           & 78.3\%          & 81.3\%          & 83.6\%          & 82.0\%          & {\ul 79.8\%}                            & {\ul 77.5\%}                            & {\ul 73.5\%}                            & 65.7\%                                  \\
BBBG         & 92.7\%          & 92.9\%          & \cellcolor[HTML]{96FFFB}\textbf{94.1\%} & \cellcolor[HTML]{96FFFB}\textbf{93.2\%} & \cellcolor[HTML]{96FFFB}\textbf{91.6\%} & \cellcolor[HTML]{96FFFB}\textbf{89.8\%} & \cellcolor[HTML]{96FFFB}\textbf{87.2\%} & \cellcolor[HTML]{96FFFB}\textbf{81.2\%} & \textbf{} & 81.3\%          & 81.3\%          & 85.2\%          & 83.3\%          & \cellcolor[HTML]{96FFFB}\textbf{83.6\%} & \cellcolor[HTML]{96FFFB}\textbf{84.0\%} & \cellcolor[HTML]{96FFFB}\textbf{76.4\%} & \cellcolor[HTML]{96FFFB}\textbf{68.6\%} \\ \hline
\end{tabular}
}
\end{table*}

\noindent {\bf BBBG outperforms prior works in scarce and biased regimes.}The best of baseline models, including deep learning methods, semi-supervised method (LS-KNN or ST-DNN), performed well when outcome labels are abundant and sampling for supervision is unbiased (others such as BERT-DNN, \citep{devlin2018bert} perform poorly (see Sec. B.1)). But with increasing scarcity their performance degrades significantly. As evident in Tab.~\ref{tab:congress} and Tab.~\ref{tab:congress2} BBBG significantly outperformed all baseline models once the supervision became lower than 20\%, and this gap widened with decreasing supervision for both biased and unbiased supervision. {With as little as 5\% labels ($\sim$10k) with biased sampling for supervision, and 3\% ($\sim$6.6k) with unbiased supervision, BBBG predictions are highly accurate in predicting party labels of authors.

\noindent {\bf GloVe vs. BERT and RoBERTa\footnote{GloVe has a natural initialization in the form of ``single-words'' that allows for estimation of contexts and context filtering. We baseline BERT and RoBERTa with SBBG because single-word initialization is difficult as their embeddings are intertwined with surroundings through attention mechanism.}}
First, note that when scarcity/bias is not an issue, BBBG with GLoVe vs. other SOTA language models, BERT and RoBERTa\citep{liu2019roberta}, perform similarly (Tab.~\ref{tab:congress}, Sec.~\ref{sec:addexpt}). However, under label scarcity (<8\%, which is about 17.6k) and selection bias, RoBERTa embedding performed no better or worse relative to GloVe when combined with either DNN or SBBG framework. This suggests that the more complex model such as RoBERTa or BERT may be more demanding on label abundance and quality and hence more vulnerable to poor supervision. In addition, the Domain Adaption (based on RoBERTa embedding) did not appear to be advantageous than some other baselines, and was significantly below BBBG under almost all conditions. Together, these results show that our BBBG model is considerably more resilient to both label scarcity and selection bias.

\begin{table*}[ht]
\vspace{-0.1in}
\caption{\footnotesize Accuracy in predicting ideology labels under unbiased/biased supervision for \textit{Gun Debate Forum} datashowing competing results between three baselines, the main model BBBG, and the single branch variant SBBG. The best results are in \textbf{bold} and the second best are \underline{underlined}. BBBG outperforms most other models substantially with scarce labels, marked in blue. The percentage shown were averaged over three independent trials.}\vspace{-0.1in}
\centering
\label{tab:forum}
\resizebox{15cm}{!}{
\begin{tabular}{lccccccccccccccccc}
\hline
             & \multicolumn{8}{c}{\textbf{Unbiased Supervsion}}                                                                                                                                                                                                                                                                                              &  & \multicolumn{8}{c}{\textbf{Biased Supervision}}                                                                                                                                                                                                                                                                       \\ \cline{2-9} \cline{11-18} 
             & 80\%                                    & 60\%                                    & 40\%                                    & 20\%                                    & 8\%                                     & 5\%                                     & 3\%                                     & 1\%                                     &  & 80\%                                    & 60\%                                    & 40\%                                    & 20\%            & 8\%                                     & 5\%                                     & 3\%                                     & 1\%                                     \\
8l-DNN       & 72.1\%                                  & 72.3\%                                  & 67.4\%                                  & 61.0\%                                  & 60.9\%                                  & 61.2\%                                  & 61.4\%                                  & 61.3\%                                  &  & 69.9\%                                  & 69.8\%                                  & 66.2\%                                  & 64.8\%          & 61.0\%                                  & 61.0\%                                  & 61.2\%                                  & {\ul 61.2\%}                            \\
LS-KNN       & 66.2\%                                  & 65.0\%                                  & 64.3\%                                  & 64.8\%                                  & 64.9\%                                  & 63.9\%                                  & 62.7\%                                  & 60.8\%                                  &  & 62.4\%                                  & 60.8\%                                  & 61.2\%                                  & 63.4\%          & 61.7\%                                  & 62.5\%                                  & 61.3\%                                  & 61.0\%                                  \\
ST-DNN       & 70.9\%                                  & 68.6\%                                  & 66.8\%                                  & 64.1\%                                  & \textless{}60.0\%                       & \textless{}60.0\%                       & \textless{}60.0\%                       & \textless{}60.0\%                       &  & 69.4\%                                  & 64.8\%                                  & 64.6\%                                  & 65.4\%          & 61.9\%                                  & 61.2\%                                  & 61.3\%                                  & \textless{}60.0\%                       \\
RoBERTa-GRU  & 76.3\%                                  & 75.7\%                                  & 75.9\%                                  & 71.0\%                                  & 63.7\%                                  & 62.2\%                                  & 61.2\%                                  & 59.8\%                                  &  & 77.6\%                                  & 73.8\%                                  & 71.5\%                                  & 68.9\%          & 67.5\%                                  & 64.2\%                                  & 61.1\%                                  & 58.1\%                                  \\
RoBERTa-DA   & 72.1\%                                  & 73.0\%                                  & 70.3\%                                  & 66.4\%                                  & 64.5\%                                  & 61.5\%                                  & 62.9\%                                  & 60.5\%                                  &  & 70.3\%                                  & 73.7\%                                  & 67.6\%                                  & 68.9\%          & 68.0\%                                  & 63.9\%                                  & 64.1\%                                  & 60.7\%                                  \\
RoBERTa-SBBG & 78.9\%                                  & 76.7\%                                  & 73.1\%                                  & 72.0\%                                  & 67.8\%                                  & 62.1\%                                  & 61.2\%                                  & 61.1\%                                  &  & 74.6\%                                  & 74.4\%                                  & 71.0\%                                  & 69.4\%          & 63.9\%                                  & 60.7\%                                  & 60.6\%                                  & 61.0\%                                  \\
SBBG         & {\ul 96.1\%}                            & {\ul 96.1\%}                            & {\ul 93.6\%}                            & {\ul 91.6\%}                            & {\ul 83.3\%}                            & {\ul 72.4\%}                            & {\ul 65.1\%}                            & {\ul 61.3\%}                            &  & {\ul 92.2\%}                            & {\ul 88.7\%}                            & {\ul 88.4\%}                            & \textbf{86.1\%} & {\ul 73.0\%}                            & {\ul 70.5\%}                            & {\ul 66.6\%}                            & 60.9\%                                  \\
BBBG         & \cellcolor[HTML]{96FFFB}\textbf{98.8\%} & \cellcolor[HTML]{96FFFB}\textbf{97.9\%} & \cellcolor[HTML]{96FFFB}\textbf{96.0\%} & \cellcolor[HTML]{96FFFB}\textbf{91.6\%} & \cellcolor[HTML]{96FFFB}\textbf{85.0\%} & \cellcolor[HTML]{96FFFB}\textbf{77.3\%} & \cellcolor[HTML]{96FFFB}\textbf{70.7\%} & \cellcolor[HTML]{96FFFB}\textbf{61.3\%} &  & \cellcolor[HTML]{96FFFB}\textbf{94.3\%} & \cellcolor[HTML]{96FFFB}\textbf{91.0\%} & \cellcolor[HTML]{96FFFB}\textbf{90.2\%} & {\ul 85.3\%}    & \cellcolor[HTML]{96FFFB}\textbf{77.3\%} & \cellcolor[HTML]{96FFFB}\textbf{74.4\%} & \cellcolor[HTML]{96FFFB}\textbf{71.5\%} & \cellcolor[HTML]{96FFFB}\textbf{61.2\%} \\ \hline
\end{tabular}
}
\vspace{-0.2in}
\end{table*}
\subsection{Gun Debate Forum Dataset}
\label{sec:gundebate}

{\bf BBBG outperforms prior works in scarce, biased and uniformly across all supervision regimes.}
Under scarcity (<8\%, about 4.8k documents) and extremity bias, we compared the performance of BBBG against several alternative models. Apart from those described above we also compared against  BERT-Sequential \citep{devlin2018bert,baly2020we}. 

BBBG significantly outperforms prior methods on the \textit{Gun Debate Forum} dataset. This is likely due to higher heterogeneity of forum users in their manner of speech compared to Congressional Dataset, which BBBG is able to handle better. We report in Tab.~\ref{tab:forum} the predicted binary ideology labels (liberal vs conservative) of posters by aggregating predicted outcomes at the author level. Excluding the SBBG ablation, RoBERTa-DA performs the best among the baselines. For SBBG, it only performs well when combining with GloVe embedding instead of RoBERTa. The gap between our model and the best baseline is widened up to 19\% difference when the supervision is biased. 

\subsection{Validation of Context Neutrality} \label{sec:neutrality}

We perform two experimental studies to illustrate the geometry of inferred latent context vectors. 

\noindent{\bf Crowd-Sourced Experiments.}
We propose to validate, using crowd-sourcing, how well words in the neighborhood of context vectors BBBG outputs aligns with human belief of neutrality. We study: 
1) Perceived relevance of words to the theme.  \\
2) To what extent are these words liberal or conservative (we also include the "do not know" option). 

To do so, we calculated the ten closest words to the inferred BBBG context vector of input documents in terms of cosine similarity in the embedded space (Sec.~\ref{sec:crowd-sourcing}, Tab.~\ref{tab:surveywords}). We then selected 6 prominent partisan themes (see Tab.~\ref{tab:surveywords},~\ref{sec:crowd-sourcing}) and for each theme randomly sampled 5 out of 10 nearest words mentioned above (hereafter, neighbourhood words). We then chose stereotypical extreme words as references for each theme. These stereotypical extreme words were manually collected by mining five liberal and conservative words/phrases from known partisan news media (see Sec. \ref{sec:crowd-sourcing} and Supplementary).
For each item, on Likert Scale, we asked gig workers to rate both relevancy (rescaled to [0,1]) and ideological leaning (rescaled to [-1,1]). 

First, we noted that both neighborhood words and manually chosen words were deemed relevant by humans. Words proximal to context vector scored above 0.843$\pm0.007$ in relevance. In comparison, the manually collected conservative and liberal reference words/phrases scored 0.703$\pm0.009$ and 0.840$\pm0.007$ respectively. On ideological leaning, the neighbourhood words scored 0.056$\pm0.018$ while the reference conservative and liberal words/phrases scored 0.323$\pm0.025$ and -0.341$\pm0.023$ respectively. The neighbourhood words were clearly more neutral (toward 0) than reference words (see Fig.~\ref{fig:sunshine}). The difference is significant at 0.001 according to a two-sample T-test. Additionally, the Spearman-rho between ranked distance from the context vector and ranked deviation from the neutral point (0) by survey was 0.5, validating that crowd-sourced ratings were aligned with ours (see Sec.~\ref{sec:crowd-sourcing}).
\begin{figure}[t]
    \centering
  \includegraphics[width=1\linewidth]{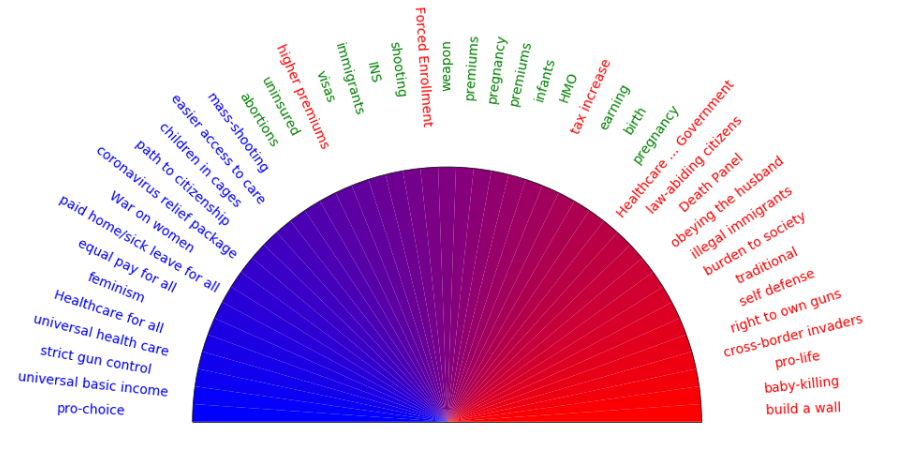} 
  \vspace{-0.3in}
        \caption{\footnotesize ``Sunshine'' plot, based on sorting crowd-sourced ratings, depicts how the crowd-workers perceive words in the neighborhood of BBBG context vector (green), relative to other stereotypical words (blue and red) appearing in partisan media. Evidently, these ratings are essentially consistent with BBBG. The few discrepancies appear to be crowd-rating noise.} \label{fig:sunshine}
        \vspace{-0.25in}
\end{figure}

\noindent {\bf Orthogonality of Latent Context and Residuals.} For each partisan theme (see Sec.~\ref{sec:themes}), $\theta$, let $\mathcal{D}_\theta/\mathcal{R}_\theta$ denote the data text or speech documents generated by Democrats/Republicans, respectively. We define the \textit{polarization axis} of $\theta$ as $\mathbf{pa_\theta}\triangleq\mathbb{E}_{x\sim\mathcal{D}_\theta}\mathbf{f}(x)-\mathbb{E}_{x\sim\mathcal{R}_\theta}\mathbf{f}(x)$, where $\mathbf{f}$ is the output of the filtered position component of BBBG. The angle between context vectors and polarization axis, averaged across different themes was about 84 degrees, which suggests near orthogonality. 

\subsection{Context Filtering leads to Purity}  \label{sec:purity}
Here we tested whether upon context filtering, the filtered position vectors, $\mathbf{f}$, are more concentrated, and exhibit better alignment with ideology labels.

We sampled the speech documents of the most frequent 20 themes given by the top 30 most active speakers (17 Democrats, 13 Republicans) from the Congress (both Senate and House included). For each document, we collected the original input text embedding $\mathbf{x}$ as well as corresponding filtered position vector $\mathbf{f}$. 
We tested two key indicators: 
(1) \% variance explained by the first principal component; 
(2) the total number of non-zero principal values above given thresholds (i.e. rank of approximated covariance matrix).

\begin{figure}[]
    \centering
    \begin{subfigure}{0.47\textwidth}
        \centering
        \includegraphics[width=0.9\textwidth]{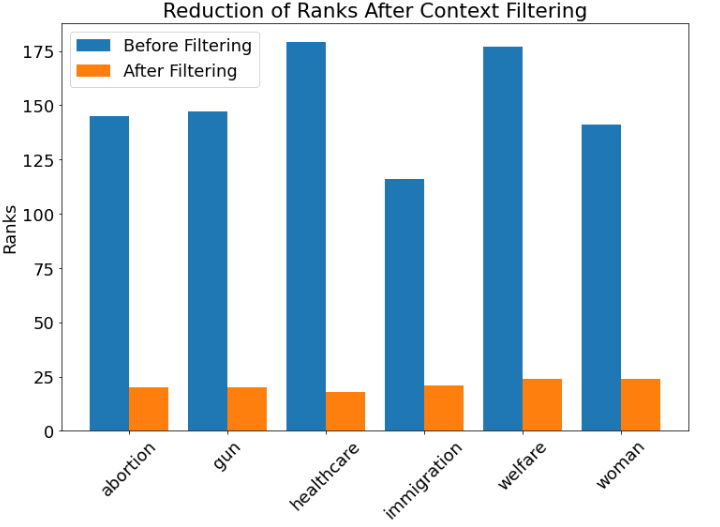}
        \end{subfigure}
    \hfill
    \begin{subfigure}{0.47\textwidth}
        \centering
        \includegraphics[width=0.9\textwidth]{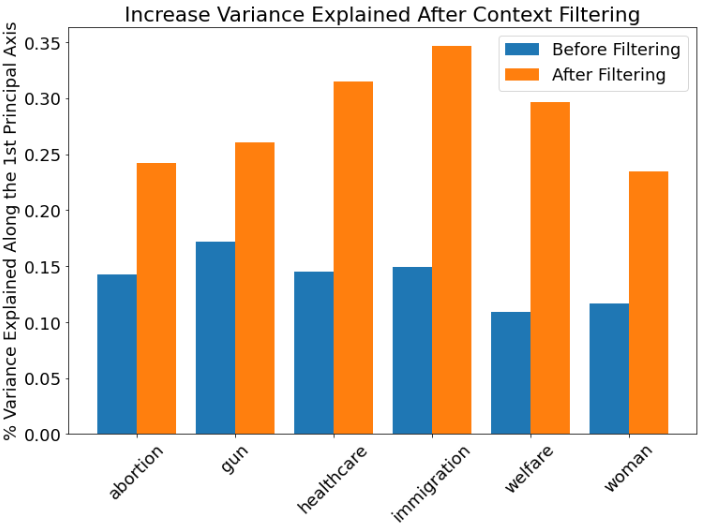}
        \vspace{-0.12in}
        \end{subfigure}
        \vspace{-0.05in}
 \caption{\footnotesize Impact of Context Filtering on Ideology Prediction. (Top) Position Vectors for each theme are concentrated in a low-dimensional space relative to unfiltered encoding. (Bottom) \% Variance explained by position vectors is higher relative to unfiltered encoding. These demonstrate significant amplification of ideological signal among a few dimensions.}
       \label{fig:expvar}
       \vspace{-0.15in}
        \end{figure}


\noindent {\bf Multiple Authors Writings on Diverse Themes.} We explore several variations to highlight de-noising effect of BBBG. We report results at a p-value of $0.001$: \\
1) {\it Author writings on one theme:} The \% variance explained along the first principal axis increased from 0.22 to 0.39 whereas the ranks decreased from 53 to 14, averaged across all authors and themes.\\
2) {\it Author writing on multiple themes:} The \% variance explained increased from 0.29 to 0.41 whereas the ranks decreased from 101 to 18, averaged across all authors.\\ 
3) {\it Multiple authors of same ideology label writing on one theme:} The \% variance explained increased from 0.20 to 0.37 for Democrats, and from 0.21 to 0.35 for Republicans, whereas the ranks decreased from 111 to 18 for Democrats, and from 147 to 31 for Republicans, averaged across themes.\\
4) {\it Multiple authors of same ideology label writing on multiple themes:} the \% variance explained increased from 0.26 to 0.40 for Democrats, and from 0.22 to 0.33 for Republicans, whereas the ranks decreased from 151 to 29 for Democrats, and from 147 to 31 for Republicans. 

\begin{figure*}[h]
    \centering
    \begin{subfigure}{0.3\textwidth}
        \centering
        \includegraphics[width=2in]{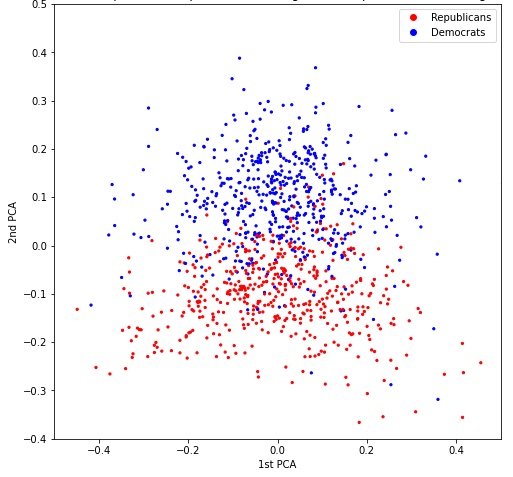}
        \caption{}\label{fig:Zbar}
    \end{subfigure}
    \hfill
    \begin{subfigure}{0.3\textwidth}
        \centering
        \includegraphics[width=2in]{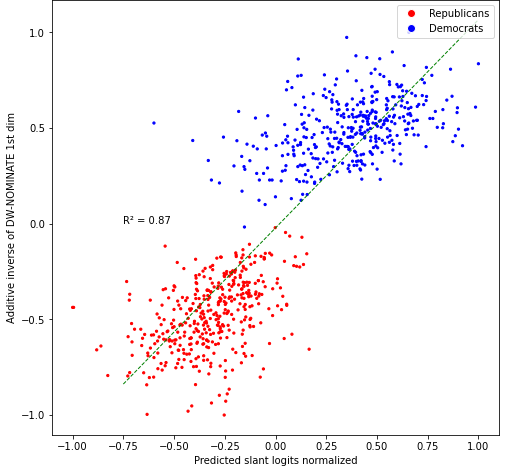}
        \caption{}\label{fig:bbbg_dw}
    \end{subfigure}
    \hfill
    \begin{subfigure}{0.3\textwidth}
        \centering
        \includegraphics[width=2in]{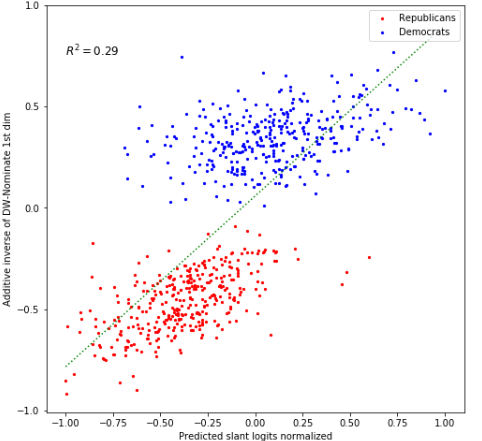}
        \caption{}\label{fig:lsknn_dw}
        \end{subfigure}
        \vspace{-0.1in}
 \caption{\footnotesize (a) Top 2 PCA components of the ideology vector, $\mathbf{z}$, suggests that it captures party affiliation. 
 (b\&c) Correlation between additive inverse of DW-NOMINATE (y-axis) and slant value (x-axis). (b) BBBG with higher $R^2 \sim 0.9$ correlation. (c) best-performing baseline ($R^2\sim0.3$) 
 }
       \label{fig:validation}
        \vspace{-0.2in}
\end{figure*}
In addition, we found that among each of the six partisan themes, $\mathbf{f}$ has significant higher \% variance explained at lower ranks (Fig.~\ref{fig:expvar}). This demonstrates significant concentration in substantially fewer directions for partisan themes. Together these results consistently suggest context filtering leads to concentration over several "key" directions, and that the inputs used to predict ideology become less "noisy." 

\noindent {\bf The filtered position vectors for different parties are better-separated than original embeddings.} 
Here we sampled the top 20 most frequent themes keeping the other aspects same. 
In previous analysis we demonstrated that the first principal axis explained at least 35\% of total variation from inputs represented as residuals, or 20\% for inputs represented as document embeddings. With context filtering the distance between ideological centers \footnote{Defined as center of normalized vectors projected onto the 1st Principal axis} of each party increased 5-fold (from 0.004 to 0.02). Furthermore, when grouped by themes, we observed heterogenous effects, for themes such as "abortion", "guns", "healthcare", and "Iran/Lybia/Syria", the distanced increased significantly (by more than 0.05), while for themes such as "culture", "religion", and "sports", the distances had marginal effect. {\it Thus themes that are more polarized were better separated along the 1st (i.e. the dominant) Principal Axis.}

\noindent
{\bf Better Ideological Alignment with Labels.}
To verify whether our model learned a correct representation of ideology, We tested on \textit{Congressional Speeches}, training with  8\% biased supervision we observed the following:\\
(1) 
Evidently, ideological members are separated along the 2nd principal component, indicating that  $\mathbf{z}$ is indeed an ideological representation (Fig \ref{fig:Zbar}).\\ (2) Fig \ref{fig:bbbg_dw} shows the mean of the logit output of the ideology supervision module (Fig \ref{fig:diagram}) aggregated by author is highly correlated ($R^2\sim0.9$) with \textit{DW-NOMINATE} scores, which is a dataset describing the ideology scores of House and Senate members based on their voting records, and considered a gold-standard (see Sec.~\ref{sec:dataset_append}). As such \textit{DW-NOMINATE} scores are agnostic to congressional speech, and a high correlation implies BBBG accurately captures ideological differences, compared to baseline models (Fig \ref{fig:lsknn_dw}).

\subsection{Knowledge Transfer to Unseen Themes} \noindent {\bf Better Transfer to Unseen Themes.} Here we train samples with supervision labels for documents belonging to eight themes ('IT', 'abortion', addictive', 'economy', 'political actions', 'postal service', 'sport', 'traditional energy', 'waste', 'workforce'), which constitute 8\% of total documents and contain both partisan and non-partisan themes. We then test on 60 "unseen" themes (see Sec.~\ref{sec:themes}. 
As shown in the Tab. \ref{tab:bytopics}, BBBG outperformed all other baselines, which demonstrates that context filtered position vectors reveal ideological similarity across different unrelated themes. 

\begin{table}[t]
\caption{\footnotesize Experiments on Congressional Reports with supervision labels coming exclusively from samples of eight chosen themes. Here we reported prediction accuracy among document samples of 60 other themes (see Sec.~\ref{sec:themes}). The best results are in bold and the second best are underlined. BBBG outperforms all other models substantially. The percentage shown were averaged over three independent trials}
\label{tab:bytopics}
\centering
\resizebox{\columnwidth}{!}{
\begin{tabular}{cccccc}
\hline
LS-kNN & \begin{tabular}[c]{@{}c@{}}RoBERTa\\ -DA\end{tabular} & RoBERTa & 8l-DNN & SBBG         & BBBG                                    \\ 
63.8\% & 50.1\%                                                & 58.2\%  & 78.8\% & {\ul 81.6\%} & \textbf{86.1\%} \\ \hline
\end{tabular}
}
\vspace{-0.2in}
\end{table}

\noindent {\bf BBBG Transfers better from Extreme to Near-Neutral.} We evaluate the efficacy of BBBG in generalizing across different biased distributions (See Appendix Fig.~\ref{fig:profile}). Here conservative or extremely conservative posters are about 45\% of the posters, and liberal or very liberal are 25\%. 

\noindent This problem is of particular relevance since the majority of the US population is ideologically non-extreme, yet politically inactive \citep{wojcik2019sizing}, and therefore important to understand \cite{delli2000gen}.
%
We evaluate our trained models (see Tab.~\ref{tab:light extreme}) on Gun Debate Forum, and on the sub-population of slightly leaning posters as well as on the held-out set of extreme posters (namely those masked in training). ST-DNN is reported because it is the most representative among prior works, and as observed BBBG's outperforms ST-DNN. Similar results are observed with other baseline models such as 8l-DNN. 
\begin{table}[t]
\vspace{-0.1in}
\caption{\footnotesize Comparing prediction accuracy of ideology labels among extreme and slightly leaning groups by models trained on different levels of biased supervision}
\vspace{-0.1in}
\centering
\label{tab:light extreme}
\resizebox{\columnwidth}{!}{
\begin{tabular}{lccccc}
\hline
                                                            & \multicolumn{2}{c}{\begin{tabular}[c]{@{}c@{}}Slightly\\ Leaning\end{tabular}} & \multicolumn{1}{l}{} & \multicolumn{2}{c}{Extreme} \\ \cline{2-3} \cline{5-6} 
\begin{tabular}[c]{@{}l@{}}Supervision\\ Level\end{tabular} & ST-DNN                                   & BBBG                               &                      & ST-DNN         & BBBG      \\ \cline{1-3} \cline{5-6} 
80\%                                                        & 69.3\%                                & 83.1\%                                 &                      & 98.6\%      & 99.2\%        \\
60\%                                                        & 63.2\%                                & 74.2\%                                 &                      & 93.4\%      & 99.1\%        \\
40\%                                                        & 58.7\%                                & 68.2\%                                 &                      & 88.4\%      & 98.5\%        \\
20\%                                                        & 59.4\%                                & 65.9\%                                 &                      & 81.2\%      & 94.5\%        \\
8\%                                                         & 58.4\%                                & 60.1\%                                 &                      & 70.4\%      & 85.5\%        \\
5\%                                                         & 50.4\%                                & 56.5\%                                 &                      & 70.6\%      & 81.9\%        \\ \hline
\end{tabular}
}
\vspace{-0.2in}
\end{table}

\vspace{-0.05in}

\section{Conclusion}
We propose a novel deep-learning method to deal with label scarcity and selection bias that typically arise in political ideology prediction problems. Our method learns to decompose text embeddings into neutral contexts and context filtered position vectors, which contain ideological information. In addition to demonstrating improved prediction accuracy on benchmark datasets, we also expose important aspects of our approach through ablative experiments. We validate context neutrality through crowd-sourcing, ideological concentration through context filtering, and knowledge-transfer to documents dealing with novel themes, and to people, whose ideology bears little similarity to training data. Going forward, we can check if our model can extend to more general social platforms such as Twitter, or learn and verify ideological representation on a continuum similar to DW-NOMINATE.

\section{Ethical Considerations}
\subsection{Data Collection and Usage}

Congressional Report corpus is publicly available and can be directly downloaded online. Posts from Debatepolitics.com were collected in a manner that strictly followed the terms of use of the original sources and the privacy rights of the source owner. Authors involved in the data collection process have all finished their human subjects research training. 

The Debatepolitics data will be released upon request. The personal information of forum posters will be concealed from the public. 

\subsection{Benefit and Potential Misuse of BBBG}

\textbf{Intended Use}.The goal of this project is to provide a means to overcome the label bias and scarcity problem that has not been fully addressed in the ideology prediction literature. It also provides a useful representation of ideology that can be further explored for other learning purposes. It is particularly useful to predict and evaluate the stance of the non-extreme group who tends to politically inactive (cf. sec 4.2.3). 

Recent Kansas abortion vote \footnote{\url{https://apnews.com/article/kansas-abortion-vote-recount\\-e874f56806a9d63b473b24580ad7ea0c}} has demonstrated the importance of predicting leanings of the silent majority. Devoid of such tools, lawmakers are more likely to incorrectly extrapolate views of the vocal minority to the entire population. Furthermore, poor extrapolation emanating from the vocal minorities views can have a significant impact on political disengagement\footnote{\url{https://academic.oup.com/pa/article/68/suppl_1/241/1403570}}.

However, like any machine learning model, there is a risk of over-generalization of its true capabilities. The output of our model needs to be assessed and evaluated with full consideration of the characteristics of the input source. The potential domain difference of authors of texts might be significant, and any conclusion drawn from studying our group of authors cannot be immediately generalized to other groups. 

\textbf{Risk of Misuse and Potential Harm}. Our model should not cause harm unless its users interpret the prediction results in an unintended way. It is meant to provide insights on the ideology distribution of a group or a population, instead of judgment of an individual. Its output is not without error, albeit more accurate than most models under realistic situations. And for slightly leaning and moderate people, it is possible our model may generate incorrect outputs relative to the ground truth. Though, our model mitigates this relative to the prior SOTA. The potential harm of our model could be magnified if it is used in making decisions on vulnerable populations. 

The predictions and insights generated by our model should not be treated as facts or golden rules. We also suggest that results from any political related studies should be interpreted with skepticism and encourage the users of our model to perform careful evaluation in their corresponding application domain, check more sources or consult political scientists for expert opinions.

\newpage
\bibliography{reference.bib}
\bibliographystyle{acl_natbib}
\appendix
\newpage
\section{Appendix}
\label{sec:appendix}

\subsection{Dataset Details.} \label{sec:dataset_append}
\noindent \textit{Congressional Speeches} is a corpus of transcriptions of (220k) speeches given by house or senate congresspeople during 2009-2020, spanning both the Obama and Trump presidencies. For each speech, the speaker ID, year, party affiliation of the speaker, and the title of the speech was provided. The ideological label of each speech document is given as the party affiliation of its speaker. Although such a proxy seems imperfect, it has been shown that party affiliations are largely consistent with congresspeople' ideology\citep{levitt1996senators}. Extremity of documents/speakers can be ascertained from DW-Nominate scores.

\noindent \textit{Gun Debate Forum} is a corpus of 60K posts on the online forum (debatepolitics.com) debating the issues of gun violence and gun control. Each user  posting in this forum may choose to affiliate with one of the following ideological groups: slightly liberal, slightly conservative, liberal, conservative, extreme liberal, extreme conservative, moderate, centrist, libertarian, anarchist, and progressive. In this study, we limited our attention to the liberal-conservative spectrum, including posts from users who identified as: slightly liberal, slightly conservative, liberal, conservative, extreme liberal, extreme conservative, and moderate (Fig.~\ref{fig:profile}). This places all posts and their authors on a 7-point scale \citep{preoctiuc2017beyond}.

\noindent \textit{DW-NOMINATE} is a dataset describing the ideology scores of House and Senate members based on their voting records, obtained from (voteview.com). Scores (primary dimension) range continuously from liberal (-1) to conservative (+1) and explain the majority (>80\%) of voting differences. Details of how scores are calculated from voting data are provided in \citep{poole2001d}. DW-NOMINATE scores are widely considered as a benchmark metric \citep{jackson1992ideology}.

\subsection{Baseline Methods and Experiment Details} \label{sec:methoddetails}
\textbf{8l-DNN} A fully-connected (FC) network that has 7 intermediate layer and an output layer of size 1. For GloVe embedding, the input document embedding is obtained by averaging word embeddings with the size of 300, while for BERT, the input document is represented using CLS (or pooled) embedding with the size of 1024. The intermediate layers are of size (in order) 800, 800, 800, 400, 250, 800, and 800 (to reduce verbosity, the shape of such FC network will be written as (800, 800, 800, 400, 250, 800, 800, 1), same for other FCs hereinafter). We used ReLU as activation function except for the output layer where we used Sigmoid function. We used 0.001 learning rate for training, l2-regularization at 0.01 at the last two layers, and RMSProp for optimization\citep{hinton2012neural}.
The 8l-DNN will henceforth be used as building blocks for some other baseline models. Unless specified, it will use the same structure and parameters as above. 

\textbf{The GS model} originated from the benchmark method measuring political slants from texts, developed by Gentzkow and Shapiro in 2020 \citep{gentzkow2010drives}. It is based on the Na\"ive Bayes assumption. We repeated what described in \cite{gentzkow2010drives} by picking out most polarized phrases, and then regressing party labels over word frequency, and then use the sum of coefficients of those polarized phrases weighted by phrase frequency to obtain the party slant of each speech in the test data. 

\textbf{GRU} This model takes texts as sequences of input embedding\citep{dey2017gate}, and output a vector of length 300. This output vector was further fed into a Dropout layer (p =0.5), then into a FC network of a shape (800, 800, 400, 1). All layers use ReLU activation except for the last layer which uses Sigmoid. To train this GRU, we used ADAM optimizer with starting learning rate of 0.01

\textbf{The BERT/RoBERTa model} Both the BERT and RoBERTa model output two types of representation for sentences - pooled/CLS embeddings or sequences of embeddings \citep{devlin2018bert,baly2020we,liu2019roberta}. We tried both in our experiments and only reported results from sequential representations from RoBERTa as it produced the best performance. The sequence of embeddings were fed as input into a network of the same structure described in the GRU. For the RoBERTa-SBBG model, the sequence of embeddings were first fed as input into a single GRU layer to generate embeddings of size 300, which were subsequently treated as input samples for a framework described in SBBG model\footnote{We only tested RoBERTa with the SBBG ablation instead of BBBG full model due to the fact that BERT/RoBERTa embeddings are intrinsically non-linear which are created from weighted sum of all contextual information, and hence conflicted from the linear decomposition model of documents which is the backbone of BBBG. However, as shown in our experiments, switching to RoBERTa from GloVe offered little benefits in improving the learning performance.}. Unlike GloVe, we allowed the embeddings to be fine-tuned during training.

\textbf{Domain Adaptation} The Domain Adaptation module was based on sequence embeddings produced by RoBERTa. During the training stage, the labeled texts were assigned to the source domain, and equal number of randomly sampled unlabeled texts were assigned to the target domain. During the inference stage, predictions from the label classifier were reported on the test data.

\textbf{Label Spreading} and \textbf{Self Training} are both prototypical semi-supervised learning (SSL) models to deal with situations in which the training data are not sufficiently labeled \citep{yarowsky1995unsupervised, ortigosa2012approaching}. Although supervision insufficiency is a predominant obstacle in ideology learning, SSL is rarely used in this field of study and was included to see why it was not widely adopted to address this issue. In our experiments, we used kNN (k=7) kernel for Label Spreading, as well as 8l-DNN, Random Forest or XGBoost as kernels for Self Training\citep{rodriguez2006rotation,chen2016xgboost}. 

\begin{figure}[t]
    \centering
  \includegraphics[width=\linewidth]{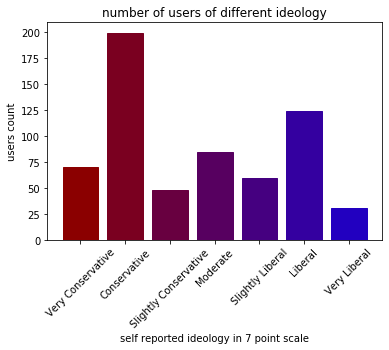} 
        \caption{\footnotesize Number of posters per ideology group. Observe that conservative or extremely conservative posters are about 45\% of the posters, and liberal or very liberal are 25\%. The situation is reversed among slightly leaning posters. They are of slightly liberal than slightly conservative.} \label{fig:profile}
        \vspace{-0.15in}
\end{figure}

\subsection{BBBG Training}
According to eq3, the generative part of BBBG model can be visualized as in fig\ref{fig:dag}, while the whole model is illustrated in fig\ref{fig:diagram}. It contains four major components, listed as followed.

\begin{figure}[t]
    \centering
  \includegraphics[width=0.6\linewidth]{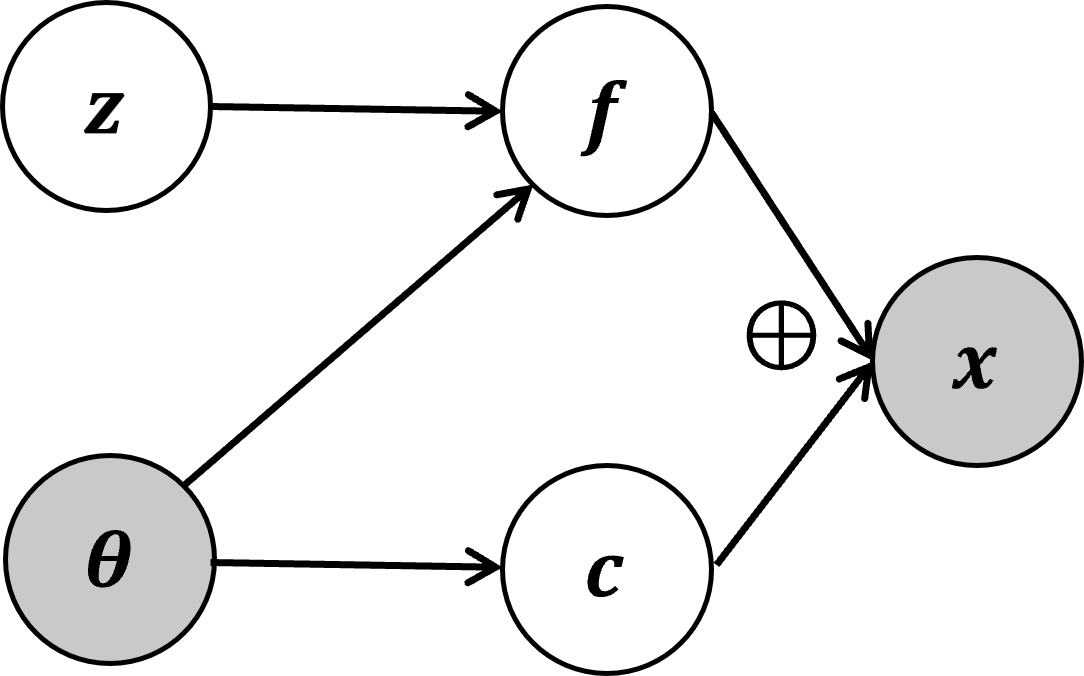} 
        \caption{A diagram of generative part of BBBG.} \label{fig:dag}
        \vspace{-0.15in}
\end{figure}

\begin{figure}[t]
     \centering
     
     \includegraphics[width=0.95\linewidth]{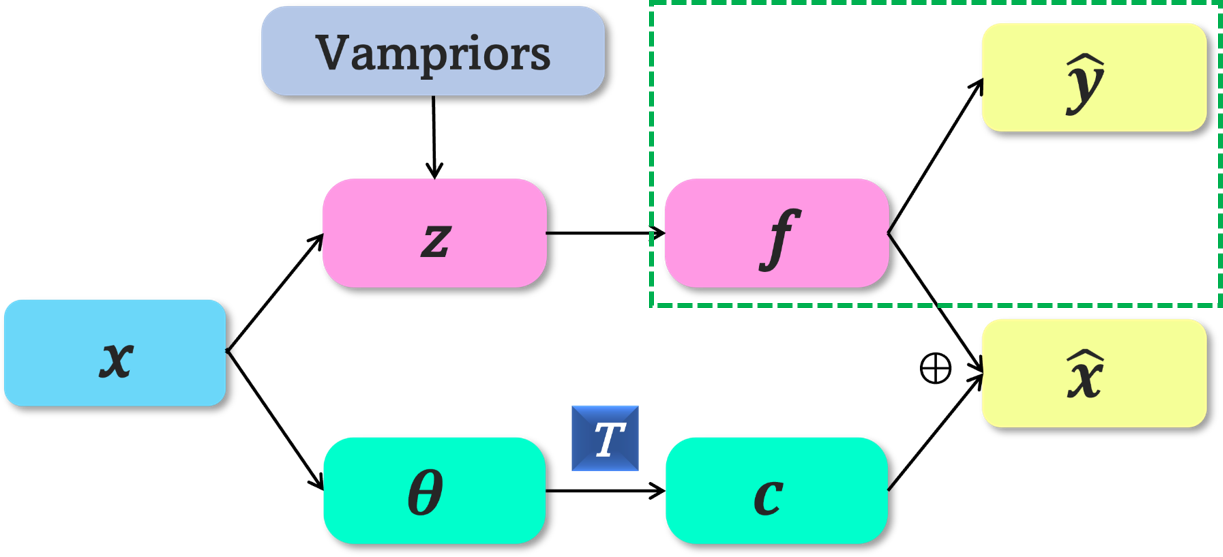}
     \caption{Proposed bi-branch BBBG network with VampPriors. Each arrow here represented multiple layers (>3), excluding the reconstruction layer. The green square highlights the ideology supervision module. 
     }
     \label{fig:diagram}
     \vspace{-0.25in}
 \end{figure}

\fbox{$~\mathbf{x}$ to $~\mathbf{z}$}. The encoder of the VAE framework, where the mean and standard deviation of $~\mathbf{z}$ (dim=50) is learned from $~\mathbf{x}$ through two fully-connected (FC) networks.$~\mathbf{z}$ is sampled through reparametrization trick. The shape of FC to learn the mean of $~\mathbf{z}$ is (800, 800, 800, 400, 50). The FC to learn the variance is of the same shape, and shares the weight with the mean network at the first two intermediate layers. All layers use ReLU activation function except for the last layer of the variance FC, which is a step function that maps 1 for all inputs between -6 to 3 but 0 elsewhere. 

\fbox{$~\mathbf{z}$ to $~\mathbf{y}$}. The ideology prediction element is composed of a FC network of a shape (800, 800, 1). It uses ReLU as the activation function, except for the last layer where it uses Sigmoid. 

\fbox{$~\mathbf{x}$ to $~\mathbf{\Tilde{\theta}}$ to $~\mathbf{c}$}. This is the neutral context branch. The theme assignment vector $~\mathbf{\Tilde{\theta}}$ is learned from x through a FC network of the shape (800, 1600, 400, 68), where 68 is the total number of themes (including "others"). The context vector $~\mathbf{c}$ is obtained by multiplying the theme matrix $T$ and $~\mathbf{\Tilde{\theta}}$ (explained below).

\fbox{$~\mathbf{z}$ and $~\mathbf{\Tilde{\theta}}$ to $~\mathbf{f}$}. The decoder of the VAE framework that decodes $~\mathbf{f}$. $~\mathbf{z}$ and $~\mathbf{\Tilde{\theta}}$ are first concatenated. Then the resulting vector will serve as the input for a FC network of a shape (800, 800, 300) to decode $~\mathbf{f}$. Next$~\mathbf{\Tilde{X}}$ will be reconstructed by sum of $~\mathbf{f}$ and $~\mathbf{c}$.

The generation of priors of $~\mathbf{z}$ (VampPriors) from pseudo-inputs follows as in \cite{tomczak2018vae}.

\textbf{Hyperparameters} For experiment on Congressional Report, we tuned hyperparameters for our main model (the single branch version inherited the same set of hyperparameters) on 10\% of total samples as the validation sets. We used 0.001 learning rate with the RMSProp Optimizer. Other parameters included the mean (0.02) and standard deviation (0.3) of Gaussian distribution where VampPriors were sampled from (the value in the bracket indicating the final tuned value)\citep{tomczak2018vae}, the annealing parameter of KL-divergence of the z prior (15)\citep{dilokthanakul2016gmvae}, the number of training epochs (10 for experiment 1, 15 for experiments 2). We used L2-regularization at 0.01 in the ideology prediction branch. Experiment on Debatepolitics corpus inherited the same hyperparameters as above.

\textbf{SBBG} Same as the BBBG but without $~\mathbf{x}$ to $~\mathbf{\Tilde{\theta}}$ to $~\mathbf{c}$, also $~\mathbf{x}$ is directly reconstructed from $~\mathbf{z}$, with a FC network of a shape (800, 800, 300).

\subsection{Initialization of Theme matrix, $T$ and theme assignment $\mathbf{\theta}$} \label{sec:initialization}
While theme assignment can be obtained by unsupervised models like the latent dirichlet allocation (LDA), we may encounter two issues. First, LDA models might uncover non-neutral associations. Second, a model purely based on word distribution might fail to uncover infrequent themes. In congressional reports, themes are dominated by fiscal spending, healthcare, economy, and politics (>60\%). Themes that were significant yet infrequent, such as guns (<1\%) and abortion (<2\%), could easily be missed by LDA models. To address these points, we adopted a popular embedding similarity approach\citep{liu2017city,sia2020tired}. First, we manually chose 67 major themes manually by digesting a significant collection of the congressional report (5K) and summaries of 5k bills; for each theme, we constructed a set of neutral words, by asking experts to come up with 5 to 10 neutral seed words (such as "firearms" and "guns" for the gun theme), and expanding into 100 words by the proximity in the word embedding space; next we manually checked each seed and expanded seed word to remove the irrelevant, rare or "non-neutral" words (i.e., words that are clearly partisan, such as "Obamacare" for the healthcare theme, "sinful" for the LGBTQ theme, and "baby-killing" for the abortion theme); then the embedding of a theme is calculated by averaging embeddings of all chosen words; lastly all theme vectors obtained above were stacked in an order to form the initial theme matrix $T_0$, of which each row corresponded to a theme vector of a certain theme, and would be subsequently used to initialize the theme vector matrix $T$ (an alternative way to decide sets of neutral words via statistical inference can be found in the next session). If we know theme assignment of a document $x$, then the context vector is $c=T\theta$. For Debatepolitics, the theme assignment $\theta$ is known. For Congressional Reports, the theme assignment was initialized by taking cosine similarity between document embeddings and context vectors obtained above \citep{liu2017city,sia2020tired}, with a slight modification. Instead of obtaining a hard assignment, we generated a initial soft assignment $\theta_0$ by taking a Softmax transformation of embedding of one document with each of the context vectors. We denoted the theme as "other" if the maximum cosine similarity falls below a certain threshold (around 10\% quantile of all maxima).

We then used these initial theme assignment in the training process, as described in the main paper, thereby allowing both $\theta$ and $T$ to be updated and fine-tuned with $x$. Shown in fig. \ref{fig:diagram}, $\hat{\theta}$ were learned from $x$ through a feedforward network, and $c$ was obtained by $c=\hat{T}\hat{\theta}$. Here $\hat{T}$ was initialized with values of $T_0$, and both $\hat{\theta}$ and $\hat{T}$ are trainable variables. In the loss function, two additional terms were included to minimize the mean-squared difference between $\hat{\theta}$ (or $\hat{T}$) and the initial values of $\theta_0$ (or $T_0$). This constrains the search for the local optima of $\theta$ and $T$ around the neighborhood of their original values.

When training was finished, we performed post hoc verification of the updated neutral context vector matrix $T$, by 1) checking whether it was orthogonal to the polarization axis corresponding to each theme (see main paper); 2) manually verified the neighborhood of the context vector for each theme in the original embedding space. For each theme, we collected the top 20 closest words to the context vectors in the embedding space. 

\subsection{An Alternative Way of Obtaining Neutral Words for the Neutral Context Vectors}

This section will offer a statistical definition of theme-wise neutral words which can be used as an alternative way to initialize the neutral context vectors. Consider a choice of theme, such as "guns", let $\mathcal{W}^D_{gun}$ be the set of words or phrases commonly used by Democrats to discuss about guns, and $\mathcal{W}^R_{gun}$ be the set of words or phrases commonly used by Republicans. For each word $\mathbf{w}\in\mathcal{W}_{gun}\triangleq \mathcal{W}^{D}_{gun}\cup\mathcal{W}^{R}_{gun}$, we calculated its document frequency $fr_{doc}$, as well as the discrepancy of document frequency among different party fraction $\Delta fr\triangleq\lvert fr^D_{doc}-fr^R_{doc}\rvert$ (or alternatively, $\chi^2$ of each words\citep{gentzkow2010drives}). Words $\mathbf{w}$ whose $fr_{doc}(\mathbf{w})\leq\alpha$ and $\Delta fr(\mathbf{w})\leq\beta$ were eliminated from $\mathcal{W}^D_{gun}$ and $\mathcal{W}^R_{gun}$. Finally the neutral set of words for the gun theme is defined as $\mathcal{W}^{\mathcal{N}}_{gun}\triangleq\mathcal{W}^D_{gun}\cap\mathcal{W}^R_{gun}$. This neutral set contains words such as "gun", "guns", "ammunition", etc. 

By this definition, words that are clearly ideologically driven (such as "libtards", "baby-killing", "death (panel)" etc.) are removed. In addition, the neutral set rules out words related to a certain theme that were preferred by a certain ideological group, such as "illegal" vs "undocumented" of the immigration theme. This allows the residual to capture as much information as possible to make the correct inference of the ideology. This approach is inspired by \citeauthor{gentzkow2010drives}'s paper, where they adopted a similar procedure, but different from us chose the most ideological words with highest $\chi^2$ values.

\subsection{Additional Details on Masking} \label{sec:addmask}

The masking procedure was performed either by random sampling or by selecting the top X\% (X ranging from 80 to 1)  of the most extreme Democrats and Republicans as unmasked. Here the extremity was determined using congress members' DW-NOMINATE score, which is considered as a benchmark metric for political ideology\citep{jackson1992ideology}. We refer to the former type of supervision as unbiased supervision (as the sample will contain both extreme and non-extreme members) and the latter type as biased supervision.

For the second experiment, the representation of texts as word embeddings and the masking procedures were the same as in experiment 1. However, due to the fact that the ground truth ideology is represented not continuously but on a 7-point scale, and the fact the distributions of users and posts from each group were uneven, we used a weighted sampling scheme \textbf{without} replacement to simulate the observed outcome scarcity of less extreme populations, as follows.

In the weighted sampling scheme, the posts generated by extreme groups (i.e. extremely liberal or conservative) were 10 times more likely to be sampled into the unmasked portion compared to the regular group (i.e. liberal or conservative), which were subsequently 20 times more likely to be sampled into the unmasked portion compared to the slightly leaning group (slightly liberal or conservative). In this way, when the unmasked samples were scarce compared to the total population (less than 8\%), they will predominantly consist of the posts generated by the extreme posters. And even when the coverage of unmasked samples reaches 80\% of the population, they are still very unlikely to include samples from the slightly leaning users. This scheme mimics the real world scenario where more extreme individuals are more politically vocal and more willing to disclose their own ideology, whereas the "silent majority", who are mostly moderate or slightly leaning, are relatively nonvocal politically, and their political views and ideology labels are largely unobserved.

 \subsection{All and Partisan Themes for Congressional Reports} \label{sec:themes}
 The names of all themes are "IP", "IT", "abortion", "academic", "agriculture", "business", "children", "China", "commerce", "crime", "culture", "homeland security", "detainee", "disadvantaged", "disaster", "race and minorities", "disease", "addictive drugs", "economy", "education", "environment", "family", "federal operation", "finance", "fiscal themes", "food", "gun", "health theme", "healthcare", "high tech", "housing", "immigration", "industry", "cyber-security", "infrastructure", "international", "Iran Syria Libya", "Iraq", "Israel", "jury", "lgbtq", "media", "military complex", "natives", "nuclear", "police", "political actions", "postal", "R\&D", "religion", "renewable energy", "reserves", "Russia", "safety", "sport", "tax", "terrorism", "trade", "traditional energy", "transportation", "veteran", "vietnam", "vote", "waste", "welfare", "woman", "workforce", and "other".
 
 The polarized themes selected to verify context vectors' neutrality (excluding uncommon themes that occurred less than 500 times) are "abortion", "agriculture", "detainee", "disadvantaged", "disaster", "race and minorities", "disease", "addictive drugs", "economy", "environment", "fiscal", "gun", "health themes", "healthcare", "immigration", "international", "Iran Syria Libya", "Iraq", "Israel", "military complex", "renewable energy", "Russia", "traditional energy", "welfare", "workforce'.
 
\subsection{Additional Details on Crowd-Sourced Experiments} \label{sec:crowd-sourcing}
To validate how well learned context vectors from BBBG aligns with human belief of neutrality, we recruited 476 participants on the Prolific platform, all of whom are citizens of the United States. Among those, the majority of them living in California (67), New York (40), and Texas (34). 50\% of the surveyees are female and 47\% are male. About 55\% of the participants have bachelor's or higher degrees, 34\% have high school diplomas, and 10\% have community college degrees. As for occupations, more than a quarter of participants report they are currently unemployed. Among the participants who are employed, most of them are working in management and professional jobs (129), followed by service (89) and sales (53). According to self-reported ideology on a seven-point scale, the majority (62\%) support Democratic viewpoints, while participants who support Republican views and those who are moderates/centrists account for 19\% each. 

We extracted context vector of six popular partisan themes (Abortion, Gun, Healthcare, Immigration, Social Welfare, and Women's Right), and calculated cosine distance (1 - cosine similarity) between context vectors and projected embedding of each of top 10K words in the lexicon. The 5 of 10 closest words for each theme was selected as \textit{context neighhorhood} words. As controls, we manually mined five stereotypical conservative/liberal words/phrases from well-established partisan news media (later we describe sources of each word/phrase). Those will be referred as \textit{reference} words. Hence, 15 words/phrases will be surveyed for each theme (75 in total). 

During the survey, each surveyee was asked to complete several demographic questions, including his/her own ideology leaning (liberal, conservative, center/neither), as well as scoring on 30 randomly sampled target or reference words/phrases. More specifically, we asked them to rate 1) to what extent (from 1 to 7 on the Likert scale) they believed those words were related to each theme, and 2) to what extent they believed those words were leaning to the liberal end or the conservative end (1 being very liberal and 7 being very conservative; an "do not know/unsure" option was also included to increase accuracy of the rating). To avoid subjective biases (for example, liberal surveyees tend to rate liberal words closer to neutral), we randomly down-sampled the results given by liberal surveyees for each words/phrases (since there are approximately twice as many liberal crowd workers as conservatives or centrists on Prolific). Eventually each word/phrase was rated by approximately the same number (around 65 on average) of liberals, conservatives, and centrists. During analysis, we re-scaled the relevancy scores to [0,1] and ideological leaning scores to [-1, 1].

We calculated the ideological leaning score for each words, and aggregated them to the theme level (see Tab.~\ref{tab:mturk}. Fig\ref{fig:abtctx} showed the ideological leaning scores of surveyed words/phrases (3 out of 5 were chosen for display clarity) for the Abortion theme. 

\begin{figure}[t]
    \centering
  \includegraphics[width=0.95\linewidth]{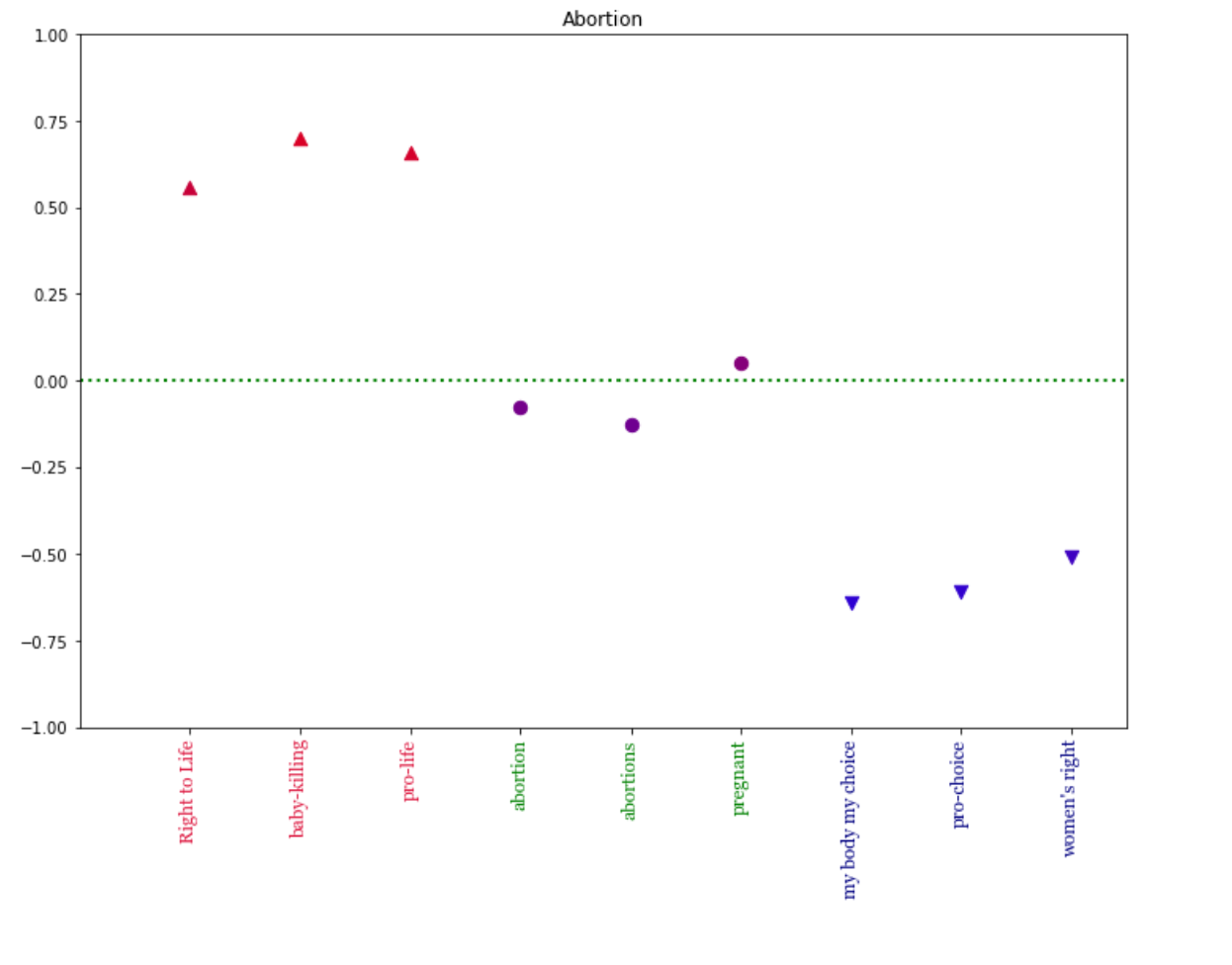} 
        \caption{Average ideological leaning scores for the Abortion theme according to crowd workers. The target, liberal, conservative words/phrases were colored in green, blue, and red respectively. y axis is ideological leaning score. The color of each point also corresponds to the value of the score.} \label{fig:abtctx}
        \vspace{-0.15in}
\end{figure}

\begin{table*}[ht]
\caption{\footnotesize Selected reference and target words/phrases for survey experiments on Prolific.}\vspace{-0.1in}
\centering
\label{tab:surveywords}
\resizebox{15cm}{!}{
\begin{tabular}{lccc}
\hline
                          & \textbf{ref\_Conservative}                                                                                                                                       & \textbf{ref\_Liberal}                                                                                                                                                                    & \textbf{Context\_Neighborhood}                                                                                                              \\ \hline
Abortion                  & \begin{tabular}[c]{@{}c@{}}Right to Life, baby-killing, immoral, \\ partial-birth abortion, pro-life\end{tabular}                                                & \begin{tabular}[c]{@{}c@{}}abortion right, pro-choice,\\ My Body My Choice, \\ reproductive right, women's right\end{tabular}                                                            & \begin{tabular}[c]{@{}c@{}}abortion, abortions, birth, \\ pregnancy, pregnant\end{tabular}                                        \\
Gun                       & \begin{tabular}[c]{@{}c@{}}2nd Amendment, gun right, \\ law-abiding citizens, self-defense\\ right to own guns\end{tabular}                                      & \begin{tabular}[c]{@{}c@{}}gun ban, gun safety, mass-shooting, \\ school massacre, strict gun control\end{tabular}                                                                       & \begin{tabular}[c]{@{}c@{}}firearms, gun, guns, \\ shooting, weapon\end{tabular}                                                  \\
Healthcare                & \begin{tabular}[c]{@{}c@{}}Death Panel, Forced Enrollment, \\ Healthcare in the hand of \\ Government Bureaucrats, \\ higher premiums, tax increase\end{tabular} & \begin{tabular}[c]{@{}c@{}}Affordable care, Healthcare for all, \\ easier access to preventive and \\ primary health care, expanding Medicaid\\ universal health care\end{tabular}       & \begin{tabular}[c]{@{}c@{}}Health Maintenance Organization, \\ healthcare coverage, Medicaid, \\ premiums, uninsured\end{tabular} \\
Immigration               & \begin{tabular}[c]{@{}c@{}}alien criminals, Build A Wall, \\ traffickers, illegal immigrants\\ cross-border invaders\end{tabular}                                & \begin{tabular}[c]{@{}c@{}}children cages, legalization of immigrants, \\ moratorium on deportations, \\ path to citizenship, undocumented immigrants\end{tabular}                       & \begin{tabular}[c]{@{}c@{}}Immigration and Naturalization Service (INS), \\ aliens, immigrants, refugees, visas\end{tabular}      \\
Welfare                   & \begin{tabular}[c]{@{}c@{}}benefit for the lazy, burden to the society, \\ opportunity society, \\ too big government, welfare queen\end{tabular}                & \begin{tabular}[c]{@{}c@{}}coronavirus relief package, \\ paid home/sick leave for all workers, \\ social safety net, universal child care, \\ Universal Basic Income (UBI)\end{tabular} & \begin{tabular}[c]{@{}c@{}}earning, incomes, premiums, \\ salary, wages\end{tabular}                                              \\
Women's Right & \begin{tabular}[c]{@{}c@{}}family values, loyalty to marriage, \\ obeying the husband, traditional, \\ virtual of women\end{tabular}                             & \begin{tabular}[c]{@{}c@{}}War on women, equal pay for equal work, \\ feminism, girls, me too\end{tabular}                                                                               & \begin{tabular}[c]{@{}c@{}}babies, infants, pregnancy, \\ teenagers, women\end{tabular}                                           \\ \hline
\end{tabular}
}
\end{table*}

\subsection{Sense-making $\mathbf{z}$: Does it Capture Ideology?}
If $\mathbf{z}$ is ideological in nature, then there should exist some principle axes along which members of common ideological groups (party labels) are differentiated. Fig \ref{fig:Zbar} shows the top two PCA components of the (mean aggregated by author) ideology $\mathbf{z}$ for all authors, with colors corresponding to author party affiliation (Republicans red; Democrats blue). With few exceptions, members are well separated along the 2nd principal component, indicating that  $\mathbf{z}$ is indeed an ideological representation.

\noindent If our model produces an accurate representation and estimation of the true ideology of authors, then the mean of output aggregated by author should be highly correlated with traditional ideology scores obtained from data independent from \textit{Congressional Speeches}, such as the \textit{DW-NOMINATE} scores based on voting behavior. Fig \ref{fig:validation} shows the relationship between our model's predicted slant score, which is the output of the machinery predicting ideological labels from $\mathbf{z}$, and that of the primary DW-NOMINATE score for authors. The two scores are highly correlated ($R^2\sim0.9$), indicating that our model accurately estimated the ideology of document authors with minimum supervision. Having established the validity of our model, we now turn to evaluations in the face of label scarcity and extremity bias.

\section{Additional Results}

\subsection{Additional Table for Experiment 1} \label{sec:addexpt}
Table \ref{tab:congress2} provides additional competing results between BBBG and several other baselines mentioned in the main paper, performed over the congressional report corpus. This is to provide evidence that methods tabulate in main paper outperform methods reported here.

\begin{table*}[t]
\caption{Accuracy of party prediction under unbiased or biased supervision for \textit{Congressional Speeches} data, showing competing results between other baselines and the main model. The best results are in \textbf{bold} and the second best are \underline{underlined}. BBBG outperforms most other models substantially with scarce labels, marked in blue. The percentage shown were averaged over three independent trials.}
\centering
\label{tab:congress2}
\resizebox{16cm}{!}{%
\begin{tabular}{lccccccccccccccccc}
\hline
         & \multicolumn{8}{c}{\textbf{Unbiased Supervision}}                                                                                                                                                                                                                                                                                             &           & \multicolumn{8}{c}{\textbf{Biased Supervision}}                                                                                                                                                                                                                                                                                               \\ \cline{2-9} \cline{11-18} 
         & 80\%                                    & 60\%                                    & 40\%                                    & 20\%                                    & 8\%                                     & 5\%                                     & 3\%                                     & 1\%                                     &           & 80\%                                    & 60\%                                    & 40\%                                    & 20\%                                    & 8\%                                     & 5\%                                     & 3\%                                     & 1\%                                     \\
ST-RF    & 70.0\%                                  & 71.5\%                                  & 71.1\%                                  & 69.7\%                                  & {\ul 69.7\%}                            & {\ul 69.8\%}                            & {\ul 72.8\%}                            & 61.1\%                                  &           & {\ul 74.3\%}                            & {\ul 72.9\%}                            & {\ul 68.9\%}                            & 56.9\%                                  & 61.3\%                                  & 61.3\%                                  & 61.3\%                                  & 61.3\%                                  \\
BERT     & 67.4\%                                  & 67.3\%                                  & 66.4\%                                  & 60.4\%                                  & 53.8\%                                  & 50.4\%                                  & 50.0\%                                  & 51.2\%                                  &           & 64.7\%                                  & 65.4\%                                  & 66.1\%                                  & 57.7\%                                  & 64.8\%                                  & 61.0\%                                  & 61.0\%                                  & 61.2\%                                  \\
RF       & 65.1\%                                  & 64.8\%                                  & 64.6\%                                  & 64.6\%                                  & 64.4\%                                  & 63.6\%                                  & 63.0\%                                  & {\ul 61.2\%}                            &           & 61.7\%                                  & 57.5\%                                  & 54.2\%                                  & 53.0\%                                  & 63.4\%                                  & 61.7\%                                  & 62.5\%                                  & 61.3\%                                  \\
GRU      & {\ul 77.7\%}                            & {\ul 77.3\%}                            & {\ul 75.7\%}                            & {\ul 72.4\%}                            & 69.2\%                                  & 67.5\%                                  & 65.7\%                                  & 60.3\%                                  & {\ul }    & 61.6\%                                  & 57.3\%                                  & 54.1\%                                  & 52.9\%                                  & {\ul 86.1\%}                            & {\ul 73.0\%}                            & {\ul 70.5\%}                            & {\ul 66.6\%}                            \\
BBBG & \cellcolor[HTML]{96FFFB}\textbf{92.7\%} & \cellcolor[HTML]{96FFFB}\textbf{92.9\%} & \cellcolor[HTML]{96FFFB}\textbf{94.0\%} & \cellcolor[HTML]{96FFFB}\textbf{93.2\%} & \cellcolor[HTML]{96FFFB}\textbf{91.6\%} & \cellcolor[HTML]{96FFFB}\textbf{89.8\%} & \cellcolor[HTML]{96FFFB}\textbf{87.2\%} & \cellcolor[HTML]{96FFFB}\textbf{81.2\%} & \textbf{} & \cellcolor[HTML]{96FFFB}\textbf{81.3\%} & \cellcolor[HTML]{96FFFB}\textbf{81.3\%} & \cellcolor[HTML]{96FFFB}\textbf{85.2\%} & \cellcolor[HTML]{96FFFB}\textbf{83.3\%} & \cellcolor[HTML]{96FFFB}\textbf{85.3\%} & \cellcolor[HTML]{96FFFB}\textbf{77.3\%} & \cellcolor[HTML]{96FFFB}\textbf{74.4\%} & \cellcolor[HTML]{96FFFB}\textbf{71.5\%} \\ \hline
\end{tabular}
}
\end{table*}

\subsection{Results from Ablation Experiments}

We tested a few ablations of our main model. The first one is the single branch model (SBBG) where the context learning branch was deleted. Results of this model on both benchmarks were shown in Tab. 1 and 2. As observed in Tab.~\ref{tab:congress} and Tab.~\ref{tab:forum}, uniformly for all datasets, under either biased or unbiased supervision, BBBG outperforms SSSG. Furthermore, the difference increases at higher scarcity. Since difference in performance in BBBG and SSSG can be attributed to the document decomposition, this implies that the decomposition into neutral context and position vectors results in improved accuracy and generalization to domain shifts.

We also tested impact of K taking values different from 2, such as 1 or 3. Notice that when K=1 the variational part of the model degraded into the ordinary VAE with unimodal Gaussian prior. Tab.~\ref{tab:ablation} showed experiment results of models combining different K values (1 or 3) with SBBG or BBBG (e.g. SBBG\_K1 is SBBG with K=1). For comparison purposes, we also included our main model BBBG, and its single branch variants SBBG, both of which has K value equaling to 2. As shown in Tab.~\ref{tab:ablation}, when K value deviates from 2, the model performance was worsened. And combining with SBBG would further worsen the performance.  

\begin{table}[t]
\caption{\footnotesize Accuracy in predicting ideology labels under biased supervision for \textit{Gun Debate Forum} data showing competing results between the main model BBBG, and its variants. The best results are in \textbf{bold} and the second best are \underline{underlined}. BBBG outperforms most of its variants substantially with scarce labels, marked in blue. The percentage shown were averaged over three independent trials.}\vspace{-0.1in}
\label{tab:ablation}
\resizebox{8cm}{!}{
\begin{tabular}{lcccccccc}
\hline
         & 80\%                                   & 60\%                                                           & 40\%                                                           & 20\%            & 8\%                                                            & 5\%                                                            & 3\%                                                            & 1\%             \\ \cline{2-9} 
SBBG\_K1 & {\ul 92.4\%}                           & {\ul 90.2\%}                                                   & 86.5\%                                                         & 81.5\%          & 73.8\%                                                         & 69.9\%                                                         & 64.7\%                                                         & 60.4\%          \\
SBBG\_K3 & 91.0\%                                 & 89.0\%                                                         & 86.9\%                                                         & 82.3\%          & 73.7\%                                                         & 67.7\%                                                         & {\ul 67.9\%}                                                   & 59.5\%          \\
BBBG\_K1 & 91.9\%                                 & 90.0\%                                                         & 86.8\%                                                         & 83.8\%          & {\ul 77.2\%}                                                   & 70.4\%                                                         & 64.1\%                                                         & 58.4\%          \\
BBBG\_K3 & 90.5\%                                 & 88.0\%                                                         & 86.8\%                                                         & 83.3\%          & 76.6\%                                                         & 64.4\%                                                         & 66.1\%                                                         & 59.2\%          \\
SBBG     & 92.2\%                                 & 88.7\%                                                         & {\ul 88.4\%}                                                   & \textbf{86.1\%} & 73.0\%                                                         & {\ul 70.5\%}                                                   & 66.6\%                                                         & 60.9\%          \\
BBBG     & {\color[HTML]{000000} \textbf{94.3\%}} & \cellcolor[HTML]{96FFFB}{\color[HTML]{000000} \textbf{91.0\%}} & \cellcolor[HTML]{96FFFB}{\color[HTML]{000000} \textbf{90.2\%}} & {\ul 85.3\%}    & \cellcolor[HTML]{96FFFB}{\color[HTML]{000000} \textbf{77.3\%}} & \cellcolor[HTML]{96FFFB}{\color[HTML]{000000} \textbf{74.4\%}} & \cellcolor[HTML]{96FFFB}{\color[HTML]{000000} \textbf{71.5\%}} & \textbf{61.2\%} \\ \hline
\end{tabular}
}
\end{table} 

\subsection{BBBG can discriminate among slightly leaning groups even when this information is absent in training data.}
BBBG and prior works report meaningful performance with sufficient supervision. Therefore, a key insight of this experiment is that in heterogenous group settings, when there is even less data for slightly leaning groups because of biased supervision, a model that performs well must be particularly capable of extracting ideological knowledge. 



We compared the different schemes using a novel rank-deviation (RD) metric (see supplementary) that compares the ranking of an author in the ideological spectrum against the median rank.
%
%
%
As evident, the average or median RDs of the slightly leaning subgroup are smaller than RDs of the extreme subgroup (see Fig. \ref{fig:spread}) 
BBBG achieves better separation between extreme and slightly leaning subgroups and lower variance, compared with the ST-DNN. 



\begin{figure}[ht]
\begin{center}
%
\includegraphics[width=\linewidth]{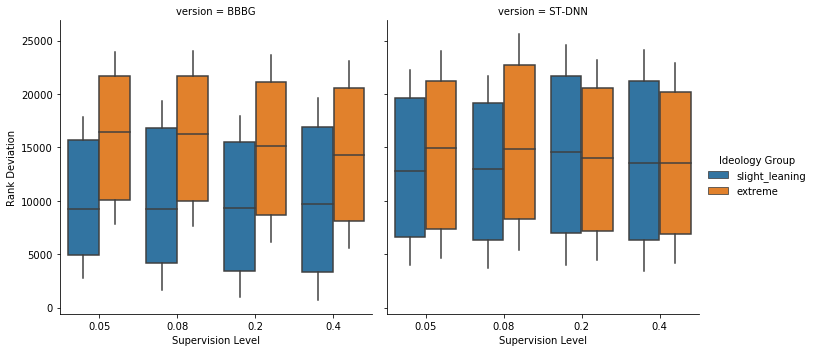}
\end{center}
\caption{Comparing the distributions of Rank Deviation of various subgroup, according to BBBG (left) and ST-DNN (rank), trained on different levels of supervision (percentage along X-axis).}
\label{fig:spread}
\vspace{-0.12in}
\end{figure}

\subsection{Additional Results on Crowd-Sourced Experiments}

Detailed comparison of crowd-rated ideological leaning score by each theme can be found in Tab.~\ref{tab:mturk}.

\begin{table}[t]
\caption{Ideological leaning of words according to crowd-sourced experiments, breaking down by themes. Each gig worker was asked to rate randomly sampled words that belong to our targeted neighborhood words or one of the reference group (Liberal words or Conservative words). Scores are re-scaled to the range of -1 to 1, where -1/1 corresponds to extreme liberal/extreme conservative leaning.}
\centering
\label{tab:mturk}
\resizebox{7cm}{!}{%
\begin{threeparttable}
\begin{tabular}{lcclc}
\hline
\multicolumn{1}{c}{} & \multicolumn{2}{c}{References}                                                                                           &  & Targets                                                      \\ \cline{2-3} \cline{5-5} 
Themes               & Conservative                                               & Liberal                                                     &  & \begin{tabular}[c]{@{}c@{}}Neighborhood\\ Words\end{tabular} \\ \cline{1-3} \cline{5-5} 
Abortion             & \begin{tabular}[c]{@{}c@{}}0.524***\\ (0.082)\end{tabular} & \begin{tabular}[c]{@{}c@{}}-0.507***\\ (0.062)\end{tabular} &  & \begin{tabular}[c]{@{}c@{}}0.039\\ (0.052)\end{tabular}      \\
Economy              & \begin{tabular}[c]{@{}c@{}}0.166\\ (0.074)\end{tabular}    & \begin{tabular}[c]{@{}c@{}}-0.341***\\ (0.073)\end{tabular} &  & \begin{tabular}[c]{@{}c@{}}0.093\\ (0.05)\end{tabular}       \\
Finance              & \begin{tabular}[c]{@{}c@{}}0.047\\ (0.073)\end{tabular}    & \begin{tabular}[c]{@{}c@{}}-0.125***\\ (0.077)\end{tabular} &  & \begin{tabular}[c]{@{}c@{}}0.058\\ (0.046)\end{tabular}      \\
Gun                  & \begin{tabular}[c]{@{}c@{}}0.498***\\ (0.067)\end{tabular} & \begin{tabular}[c]{@{}c@{}}-0.338***\\ (0.079)\end{tabular} &  & \begin{tabular}[c]{@{}c@{}}0.157\\ (0.067)\end{tabular}      \\
Healthcare           & \begin{tabular}[c]{@{}c@{}}0.097*\\ (0.081)\end{tabular}   & \begin{tabular}[c]{@{}c@{}}-0.413***\\ (0.06)\end{tabular}  &  & \begin{tabular}[c]{@{}c@{}}-0.0\\ (0.052)\end{tabular}       \\
Immigration          & \begin{tabular}[c]{@{}c@{}}0.514***\\ (0.077)\end{tabular} & \begin{tabular}[c]{@{}c@{}}-0.223**\\ (0.083)\end{tabular}  &  & \begin{tabular}[c]{@{}c@{}}-0.041\\ (0.069)\end{tabular}     \\
Iraq                 & \begin{tabular}[c]{@{}c@{}}0.317***\\ (0.063)\end{tabular} & \begin{tabular}[c]{@{}c@{}}-0.126***\\ (0.071)\end{tabular} &  & \begin{tabular}[c]{@{}c@{}}0.108\\ (0.055)\end{tabular}      \\
Renewable            & \begin{tabular}[c]{@{}c@{}}0.246***\\ (0.074)\end{tabular} & \begin{tabular}[c]{@{}c@{}}-0.439***\\ (0.053)\end{tabular} &  & \begin{tabular}[c]{@{}c@{}}-0.038\\ (0.05)\end{tabular}      \\
Welfare              & \begin{tabular}[c]{@{}c@{}}0.377***\\ (0.094)\end{tabular} & \begin{tabular}[c]{@{}c@{}}-0.493***\\ (0.055)\end{tabular} &  & \begin{tabular}[c]{@{}c@{}}0.124\\ (0.047)\end{tabular}      \\
Woman                & \begin{tabular}[c]{@{}c@{}}0.418***\\ (0.07)\end{tabular}  & \begin{tabular}[c]{@{}c@{}}-0.364***\\ (0.074)\end{tabular} &  & \begin{tabular}[c]{@{}c@{}}0.08\\ (0.049)\end{tabular}       \\ \cline{1-3} \cline{5-5} 
\end{tabular}
    \begin{tablenotes}
      \small
      \item The significance of crowd rated ideological leaning difference between our target words and either one of the reference group (Liberal or Conservative) was calculated via Two sample T-Test. The significance level was indicated behind reference groups.
      \item * p<0.05; ** p<0.01; *** p<0.001
    \end{tablenotes}
\end{threeparttable}
}

\end{table}

\end{document}